\documentclass[11pt]{article}

\usepackage[preprint]{acl}

\usepackage{times}
\usepackage{latexsym}

\usepackage[T1]{fontenc}

\usepackage[utf8]{inputenc}

\usepackage{microtype}

\usepackage{inconsolata}
\usepackage{todonotes}
\usepackage{tabularx}
\usepackage{booktabs} 
\usepackage{multirow}
\usepackage{amsmath}
\usepackage{amssymb}
\usepackage{graphicx}
\usepackage{subcaption}
\usepackage{fge}
\usepackage{colortbl}
\newcommand{\mysetminus}{\mathbin{\fgebackslash}}
\usepackage[framemethod=TikZ]{mdframed} 
\usepackage{listings}
\usepackage{longtable}
\usepackage{tcolorbox}
\definecolor{LightCyan}{rgb}{0.88,1,1}
\definecolor{darkgreen}{RGB}{0,150,0}

\newcommand{\rparagraph}[1]{\vspace{1.2mm}\noindent\textbf{#1.}}

\newcommand{\eg}{\textit{e}.\textit{g}.,\ }
\title{\includegraphics[width=1em]{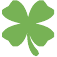}\textsc{Luq}: Long-text Uncertainty Quantification for LLMs}

\author{
Caiqi Zhang$^{1}$, 
Fangyu Liu$^{1}$\thanks{Now at Google DeepMind.}, 
Marco Basaldella$^{2}\thanks{Work done outside of Amazon.}$,
Nigel Collier$^{1}$
\\
$^1$Language Technology Lab, University of Cambridge
$^2$Amazon Alexa \\
\texttt{\{cz391, fl399, nhc30\}@cam.ac.uk} \\
\texttt{mbbasald@amazon.co.uk}
}

\usepackage{xcolor}

\definecolor{mbc}{RGB}{255, 40, 40}

\begin{document}
\maketitle
\begin{abstract}
Large Language Models (LLMs) have demonstrated remarkable capability in a variety of NLP tasks. However, LLMs are also prone to generate nonfactual content. Uncertainty Quantification (UQ) is pivotal in enhancing our understanding of a model's confidence on its generation, thereby aiding in the mitigation of nonfactual outputs. Existing research on UQ predominantly targets short text generation, typically yielding brief, word-limited responses. However, real-world applications frequently necessitate much longer responses. Our study first highlights the limitations of current UQ methods in handling long text generation. We then introduce \textsc{Luq} with its two variations: \textsc{Luq-Atomic} and \textsc{Luq-Pair}, a series of novel sampling-based UQ approaches specifically designed for long text. Our findings reveal that \textsc{Luq} outperforms existing baseline methods in correlating with the model's factuality scores (negative coefficient of -0.85 observed for Gemini Pro). To further improve the factuality of LLM responses, we propose \textsc{Luq-Ensemble}, a method that ensembles responses from multiple models and selects the response with the lowest uncertainty. The ensembling method greatly improves the response factuality upon the best standalone LLM.\footnote{\url{https://github.com/caiqizh/LUQ}}


\end{abstract}

\section{Introduction}

Large Language Models (LLMs) have demonstrated significant prowess across a wide range of NLP tasks and are increasingly being used in various downstream applications \citep{LLMSurvey, chang2023survey}. However, existing LLMs are susceptible to hallucination, often resulting in the generation of nonfactual or fabricated content \citep{manakul-etal-2023-selfcheckgpt, zhang2023siren}. One way to predict the factuality of an LLM's output without resorting to resource-intensive fact-checking procedures is by examining its uncertainty over a user query. Moreover, accurate measurement of a model's confidence in its generated responses can enable the rejection of answers with high uncertainty, potentially reducing hallucinations and improving the factuality of the output \citep{geng2023survey, wang2023survey}. 

\begin{figure*}[t!]
    \centering
    \includegraphics[width=0.99\textwidth]{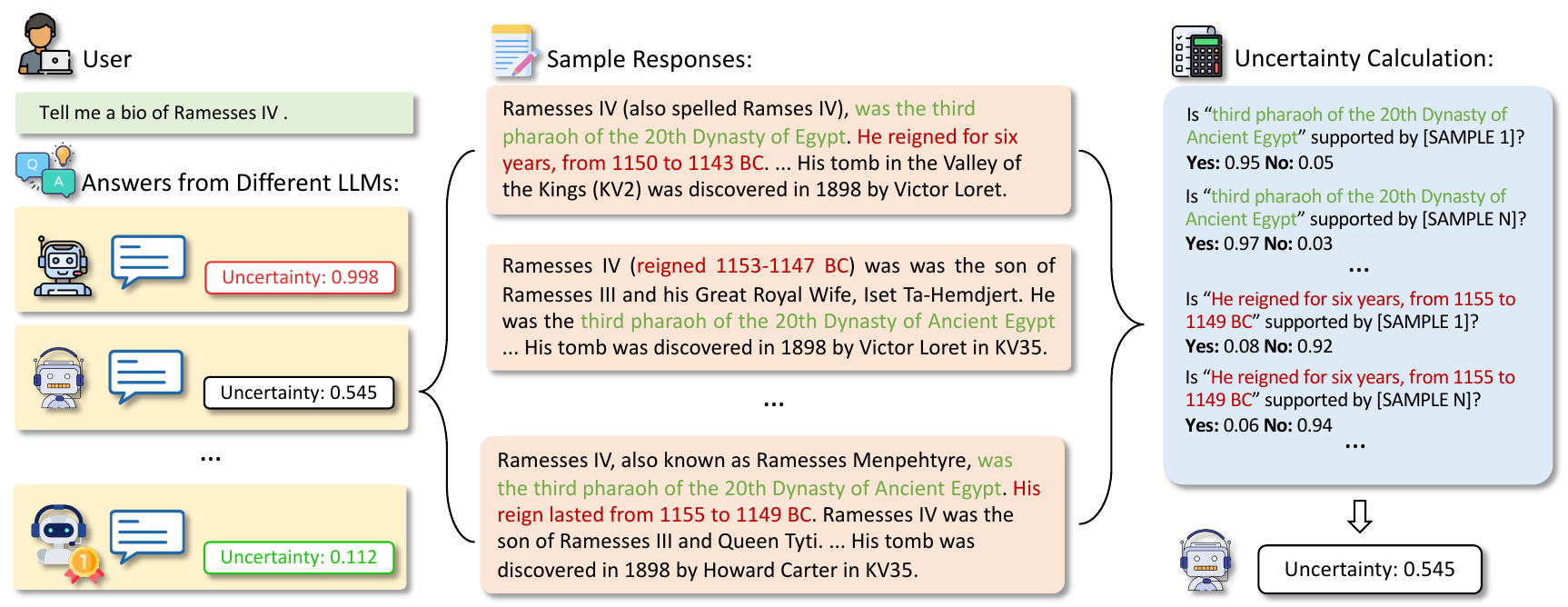}
    \vspace{-2mm}
    \caption{The illustration of the \textsc{Luq} and \textsc{Luq-Ensemble} framework. Given a question, various LLMs exhibit differing levels of uncertainty. We generate $n$ sample responses from each LLM and then assess the uncertainty based on the diversity of these samples (the \textsc{Luq} metric). \textcolor{darkgreen}{Green} highlights indicate consistency across responses (low uncertainty) and \textcolor{red}{red} highlights discrepancies (high uncertainty).  
    The \textsc{Luq-Ensemble} method selects the response from the LLM with the lowest uncertainty score as the final answer.}
    \label{fig:short_long_example}
    \vspace{-2mm}
\end{figure*}

Although Uncertainty Quantification (UQ) is a well-researched area in machine learning \citep{gawlikowski2023survey}, its application in the context of LLMs remains under-explored. One primary limitation is that previous studies on UQ mostly require access to a model's internal states (\eg logits) \citep{murray-chiang-2018-correcting, kuhn2022semantic, vazhentsev-etal-2023-efficient, duan2023shifting}. However, many best-performing LLMs, such as GPT-4 \citep{gpt4}, Gemini 1.0 Pro \citep{Gemini2023}, and Claude 2.1 \citep{anthropic2023claude}, are closed-source and only accessible via API calls. This limits the ability to directly analyze their internal processes. Another challenge is that existing research on modeling uncertainty predominantly focuses on short responses, typically less than 10 words in length \citep{kuhn2022semantic, duan2023shifting, lin2023generating}. This is in stark contrast to the more common use cases of LLMs, where responses to queries often far exceed this length, sometimes reaching hundreds of words. Such disparity points to a need for new UQ methods tailored for long-form text generated by LLMs.
Therefore, in this study we aim to answer the following research questions: \textbf{RQ1:} Are existing UQ methods still effective in the context of long-text generation? \textbf{RQ2:} If not, how can we effectively quantify LLMs' uncertainty for long-form answers? \textbf{RQ3:} In what ways can uncertainty scores be utilized to enhance the factuality of model outputs?

We explore UQ for long-text generation (at least 100 words), with an emphasis on using factuality as the key metric of the models' performance. The main contributions of this paper are:

\begin{itemize}
    \item We first highlight the limitations of existing UQ methods for long text generation and then propose \textsc{Luq} (\textbf{L}ong-text \textbf{U}ncertainty \textbf{Q}uantification; pronounced as \includegraphics[width=0.7em]{fig/four-leaf-clover.pdf}\textit{luck}), a novel UQ method that computes sentence-level consistency in long text scenarios.
    
    \vspace{-2mm}
    \item Through extensive experiments on the original \textsc{FActScore} dataset and our newly proposed \textsc{FActScore-Dis} dataset in medical domain, we demonstrate that \textsc{Luq} consistently shows strong negative correlations with the responses' factuality over 6 popular LLMs, outperforming all the baseline methods. 

    \vspace{-2mm}
    \item We propose an ensemble modeling approach that selects responses from the model exhibiting the lowest \textsc{Luq} uncertainty score, observing an improvement of up to 5\% in the overall factuality scores. Additionally, we enhance the model's uncertainty awareness by implementing a selective answering strategy.
    
\end{itemize}

\section{Background} \label{sec:background}

\subsection{Uncertainty and Confidence}

Confidence and uncertainty in the context of machine learning models pertain to the level of assurance or certainty associated with a prediction or decision \citep{geng2023survey}. While many studies treat \textit{confidence} and \textit{uncertainty}  as antonyms and use them interchangeably \citep{xiao-etal-2022-uncertainty, chen2023quantifying}, \citet{lin2023generating} provide a clear distinction: uncertainty denotes the dispersion of potential predictions for a given input, whereas confidence pertains to the degree of confidence in a specific prediction or output. We will adopt this terminology in the following sections.

Currently, a formal and universally accepted definition of uncertainty levels in language generation tasks remains elusive. Common practice in existing literature measures uncertainty through the entropy of predictions, akin to approaches in classification tasks \citep{kuhn2022semantic, lin2023generating}. Predictive entropy is formally expressed as: $$
H(Y \mid x)=-\int p(y \mid x) \log (p(y \mid x)) dy
$$

It captures the uncertainty associated with a prediction for a given input $x$. In the context of NLG, where $\mathbf{R}$ denotes all possible generations and $\mathbf{r}$ is a specific response, the uncertainty score can thus be conceptualized as:
$$
U(x)=H(\mathbf{R}\mid x)=-\sum_{\mathbf{r}} p(\mathbf{r} \mid x) \log (p(\mathbf{r} \mid x))
$$

In classification tasks, confidence for a specific prediction $y$ is quantified using the predicted probability, represented as $\hat{p}(Y=y \mid x)$ \citep{geifman2017selective, hendrycks2016baseline}. Similarly, in the context of NLG, the confidence score for a given response $\mathbf{r}$ is represented by the joint probability of the tokens in the response:
$$
C(x, \mathbf{r})=\hat{p}(\mathbf{r} \mid x)=\prod_i \hat{p}\left(r_i \mid r_{<i}, x\right) .
$$

\subsection{Uncertainty for Long Text Generation}

 
In our study, we adopt a more flexible approach to defining uncertainty and confidence in long text generation. Similar to \citet{huang2024uncertainty}, we focus on the ability of UQ methods to effectively rank responses, differentiating between correct and incorrect predictions. This approach also aligns with the concept of \textit{relative confidence} as discussed by \citet{geng2023survey}. 
Our objective diverges from the orthogonal research direction about models' calibration, which requires models to precisely reflect their true accuracy in practical scenarios \citep{lin2023generating}.
We argue that while short-answer questions may be straightforwardly assessed using metrics such as accuracy or exact match, these standards are often unrealistic for long text generation, given the complexities of real-life probabilities.

From a practical perspective, we aim for the uncertainty score to serve as a reliable indicator of the model's  performance. This performance encompasses several dimensions of generation quality, including factuality, coherence, and creativity. Our study prioritizes factuality and the truthfulness of responses, adopting these as our primary metrics. The factuality of the responses $\mathbf{R}$ given a specific query $x$ is denoted as $F\left(\mathbf{R} \mid \mathbf{x}\right)$.
Considering two inputs $x_i$ and $x_j$, we explore the relationship between the model's uncertainty, denoted as $\operatorname{U}\left(\mathbf{x}\right)$, and the factuality. Our goal is to have:
$$
\operatorname{U}\left(\mathbf{x}_i\right)  \leq \operatorname{U}\left(\mathbf{x}_j\right) 
\Longleftrightarrow F\left(\mathbf{R} \mid \mathbf{x}_i\right)  \geq F\left(\mathbf{R} \mid \mathbf{x}_j\right)
$$
Correspondingly, for a given input $x$, the model's confidence in generating a specific response $r$ is represented as $\operatorname{C}\left(\mathbf{x}, \mathbf{r}\right)$. Thus, we aim to establish the following relationship:
$$
\operatorname{C}\left(\mathbf{x}, \mathbf{r}_i\right)  \leq \operatorname{C}\left(\mathbf{x}, \mathbf{r}_j\right) 
\Longleftrightarrow F\left(\mathbf{r}_i \mid \mathbf{x}\right)  \leq F\left(\mathbf{r}_j \mid \mathbf{x}\right)
$$

\section{\textsc{Luq}}

In this section, we introduce our \textsc{Luq} method and its two variations (\textsc{Luq-Atomic} and \textsc{Luq-Pair}) to estimate uncertainty in long text generation. The overall framework is illustrated in Figure \ref{fig:short_long_example}. 


\rparagraph{Motivation}
Our underlying assumption posits that the greater the model's uncertainty regarding a given question $x$, the more diverse its responses to question $x$ will be. 
For instance, as shown in Figure \ref{fig:short_long_example}, the term ``\textit{third pharaoh of the 20th Dynasty of Egypt}'' is frequently supported by other sample responses, indicating the model's high confidence in this information. However, the samples suggest different reign periods for Ramesses IV; the inconsistency shows the model's higher uncertainty. 

Following the generation of $n$ responses, traditional UQ methods for short text commonly calculate the pairwise similarity among the responses
\citep{kuhn2022semantic, lin2023generating}. These pairwise similarity scores indicate the consistency between a pair of responses and play a vital role in subsequent uncertainty estimation.
However, answers to certain questions such as \textit{``Give me an introduction of ...''} and \textit{``Tell me something about ...''} may extend to hundreds of words. Longer text leads to an unexpected high similarity across all response pairs when applying previous methods. To address this issue and achieve a more nuanced similarity assessment, we propose the \textsc{Luq} uncertainty measurement with sentence-level similarity computation. Inspired by the hallucination detection method in \citet{manakul-etal-2023-selfcheckgpt}, we split each response to sentences, and check whether each sentence can be supported by other samples. 

\rparagraph{Notation}
Let $r_a$ represent the response generated by a LLM to a user query $x$. We generate an additional $n$ stochastic LLM sample responses $R = \{r_1, r_2, \ldots, r_n\}$ using the same query. The set $R' = \{r_a, r_1, r_2, \ldots, r_n\}$ encompasses all outputs from the model. 

For any given response $r_i \in R'$, the first objective is to determine how often it is supported (or entailed) by other samples. To this end, we employ an NLI classifier to assess the similarity between $r_i$ and each $r' \in R' \setminus \{r_i\}$. The output from an NLI classifier normally includes classifications of entailment, neutral, and contradiction, along with their respective logit values. It is important to note that we focus exclusively on the ``entailment'' and ``contradiction'' classes, as sentences labeled as ``neutral'' generally do not impact the overall factuality of a response. We calculate the NLI score for each sentence $s_j$ within a response $r$, and then average these scores. Formally, the similarity score $\mathcal{S}(r_i, r')$ between $r_i$ and $r'$ is defined as:
\begin{align*}
    \mathcal{P}\left(\text { entail } \mid s_j, r'\right)=\frac{\exp \left(l_e\right)}{\exp \left(l_e\right)+\exp \left(l_c\right)} \\
    \mathcal{S}(r_i, r')=\frac{1}{n} \sum_{j=1}^n P\left(\text { entail } \mid s_j, r'\right)
\end{align*}
where $l_e$ and $l_c$ are the logits of the ``entailment'' and ``contradiction'' classes, respectively. We opt to calculate $\mathcal{P}(\text{entail} \mid s_j, r')$ over $\mathcal{P}(\text{contradict} \mid s_j, r')$ because non-contradictory responses can still be largely irrelevant, indicating higher uncertainty \citep{lin2023generating}. The model's confidence in response $r_i$ and the overall uncertainty is therefore defined as:
\begin{align*}
&C(x, r_i)=\frac{1}{n} \sum\nolimits_{r' \in R' \mysetminus \{r'\}}^{} \mathcal{S}(r_i, r') \\
&U(x)=\frac{1}{n+1} \sum\nolimits_{r_i \in R'}^{} (1 - \mathcal{C}(x, r_i))
\end{align*}

Unlike \citet{kuhn2022semantic}'s method of applying an off-the-shelf DeBERTa model, we apply the DeBERTa-v3-large model \citep{he2022debertav3}, fine-tuned on the MultiNLI \citep{multinli} dataset. This choice is due to our input being a concatenation of short hypothesis (sentence $s$) and a comparatively longer premises (reference response $r'$). The format of our input aligns with the task in MultiNLI dataset, ensuring an effective assessment of consistency among the responses. 

\rparagraph{\textsc{Luq-Atomic}} To check the consistency of the generated responses in a more fine-grained manner, we implement \textsc{Luq-Atomic}, a variation of the original \textsc{Luq}. The key difference is that it first uses ChatGPT to break a response $r$ into atomic fact pieces $\{a_1, a_2, ..., a_j\}$. \textsc{Luq-Atomic} then calculates the uncertainty scores bases on atomic fact piece level ($a_j$) instead of sentence level ($s_j$). 

\rparagraph{\textsc{Luq-Pair}} The performance of our NLI classifier may be constrained by the length of the premises and hypotheses. To address this, we propose \textsc{Luq-Pair} to calculate the entailment score $s_j$ for each sentence $s'_j$ in $r'$ and select the maximum value. Formally, we define this as:
\begin{align*}
\mathcal{P}\left(\text{entail} \mid s_j, r'\right) = \max_{s'_j \in r'} \left| \mathcal{P}\left(\text{entail} \mid s_j, s'_j\right) \right|
\end{align*}

We discuss more about \textsc{Luq-Atomic} and \textsc{Luq-Pair} in Appendix \ref{app:variations}.

\section{Experiments}

\begin{table*}[ht!]
\footnotesize
\centering
\scalebox{.93}{
\begin{tabular}{l|l|*{11}{>{\centering\arraybackslash}p{0.85cm}}}
\toprule
\multicolumn{2}{c}{} & \multicolumn{3}{c}{\textbf{White-Box Methods}} & \multicolumn{7}{c}{\textbf{Black-Box Methods}} \\
\cmidrule(lr){3-5} \cmidrule(lr){6-12}
\multicolumn{2}{c}{} & MSP & MCSE & SE & LexSim & Ecc & NumSets & EigV & Deg & SCN & \textsc{Luq} \\

\hline

\rowcolor{gray!30}
\multicolumn{12}{l}{\rule{0pt}{10pt}\textsc{\textbf{FActScore-Bio}}} \\

\multirow{2}{*}{GPT-4} & PCC & - & - & - & -45.2 & -24.8 & -8.24 & -36.9 & -3.78 & -53.1 & \textbf{-60.4} \\
& SCC & - & - & - &  -36.0 & -12.7 & 4.18 & -18.7 & 6.73 & -41.8 & \textbf{-45.3} \\
\midrule
\multirow{2}{*}{GPT-3.5} & PCC & - & - & - &  -67.8 & -10.6 & -11.9 & -30.3 & -22.4 & -65.1 & \textbf{-71.3} \\
& SCC & - & - & - & -52.4 & -26.5 & -17.0 & -34.6 & -22.9 & -61.1 & \textbf{-66.6} \\
\midrule
\multirow{2}{*}{Gemini 1.0 Pro} & PCC & - & - & - & -67.2 & -50.3 & -53.0 & -72.7 & -64.4 & -84.5 & \textbf{-85.1} \\
& SCC & - & - & - &  -63.7 & -57.8 & -57.0 & -69.7 & -67.7 & \textbf{-82.4} & -81.3 \\
\midrule
\multirow{2}{*}{Yi-34B-Chat} & PCC & -20.7 & -43.9 & -55.8 & -70.1 & -27.6 & -25.7 & -49.0 & -39.8 & -70.3 & \textbf{-73.8} \\
& SCC & -22.1 & -44.3 & -53.6 & -68.2 & -45.0 & -31.3 & -51.1 & -38.9 & -72.7 & \textbf{-74.6} \\
\midrule
\multirow{2}{*}{Tulu-2-70B} & PCC & -16.8 & -32.4 & -50.5 & -55.7 & -2.13 & -20.7 & -50.1 & -53.4 & -75.6 & \textbf{-77.6} \\
& SCC & -15.4 & -34.8 & -52.7 & -61.8 & 10.1 & -18.1 & -50.3 & -54.0 & \textbf{-76.9} & -75.4 \\
\midrule
\multirow{2}{*}{Vicuna-33B} & PCC & -28.5 & -36.8 & -58.6 &  -38.3 & -18.7 & -20.0 & -60.5 & -58.3 & -66.8 & \textbf{-71.8} \\
& SCC & -27.9 & -37.4 & -57.2 & -50.6 & -14.0 & -16.6 & -61.7 & -62.4 & -66.5 & \textbf{-70.8} \\

\hline
\rowcolor{gray!30}
\multicolumn{12}{l}{\rule{0pt}{10pt}\textsc{\textbf{FActScore-Dis}}} \\

\multirow{2}{*}{GPT-3.5} & PCC & - & - & - & -41.8 & -27.9 & -7.81 & -38.8 & -13.5 & -59.0 & \textbf{-67.3}\\
& SCC & - & - & - & -39.4 & -26.0 & -6.94 & -36.9 & -16.3 & -59.1 & \textbf{-65.3}\\
\midrule
\multirow{2}{*}{Yi-34B-Chat} & PCC & -20.3 & -35.4 & -52.6 & -63.6 & -19.3 & -11.2 & -40.6 & -26.5 & -65.1 & \textbf{-70.5}\\
& SCC & -21.7 & -33.8 & -54.9 & -58.7 & -21.5 & -16.3 & -38.4 & -22.1 & -67.8 & \textbf{-72.4}\\

\bottomrule
\end{tabular}
}
\caption{Pearson and Spearman correlation coefficients (expressed as percentages) between different LLMs and various UQ methods on the FactScore dataset. We use the original factuality scores instead of the penalized ones.}
\label{tab:main_results}
\vspace{-3mm}
\end{table*}

\subsection{Dataset, Metric, and LLM Selection} \label{sec:dataset}

\rparagraph{Dataset}
When selecting the dataset, we considered three main criteria: (1) The dataset should be a long-form QA dataset. (2) There should be a well-designed and widely-accepted automatic evaluation tool. (3) The questions should be clear, specific, and have definite answers for objective evaluation.
We therefore employ \textsc{FActScore} \citep{min-etal-2023-factscore} to evaluate the factuality of our generated text. It offers automated assessment with a low error rate (below 2\%), enabling scalable application to diverse LLMs without requiring manual annotation. To supplement the extensive reliability testing of \textsc{FActScore} conducted by its creators, we performed a smaller-scale human annotation study. Our findings demonstrate a strong Pearson correlation of 0.88 between \textsc{FActScore} ratings and human factuality judgments, confirming it being a reliable reference for factuality. Please refer to Appendix \ref{app:dataset_selection} for more information about the discussion of this dataset and our validation process.

The original \textsc{FActScore} dataset (denoted as \textsc{FActScore-Bio}) includes 500 individuals' biographies from Wikidata with corresponding Wikipedia entries. To evaluate the applicability of UQ methods across different domains, we additionally developed a dataset, \textsc{FActScore-Dis}, focusing on disease entities. Details of this dataset can be found in Appendix \ref{app:factscore-dis}. 

\rparagraph{Metrics}
For each generated response, \textsc{FActScore} calculates a factuality score (FS). 
We apply \textsc{FActScore} for the first generated response ($r_a$). As the LLMs may sometime refuse to answer certain questions, to have a fair comparison, we introduce a penalized factuality score (PFS) and penalized uncertainty score (PUS). To calculate PFS and PUS, we assign a factuality score of zero and uncertainty score of one to questions that models opt not to answer. 

We then proceed to calculate both the Pearson Correlation Coefficient (PCC) and Spearman Correlation Coefficient (SCC) between the factuality scores and uncertainty scores. 
Following the criteria proposed by \citet{schober2018correlation}, we classify the correlation coefficients into five categories based on their absolute values: over 0.9 indicates a very strong correlation; 0.7 to 0.9 signifies strong; 0.5 to 0.7 suggests moderate; 0.3 to 0.5 denotes weak; 0.1 to 0.3 implies very weak; and below 0.1 means negligible correlation.

\rparagraph{LLMs} We selected six top-performing LLMs from the Arena Leaderboard \citep{zheng2023judging} for our experiments. Within our access rights, we chose three closed-sourced models: GPT-4 \citep{gpt4}, GPT-3.5 \citep{gpt3.5}, and Gemini 1.0 Pro \citep{Gemini2023}; and three open-sourced models: Yi-34B-Chat \citep{yi34bchat}, Tulu-2-70B \citep{ivison2023camels}, and Vicuna-33B \citep{zheng2023judging}. For each LLM, we include the following baseline UQ methods for comparison. Our implementation is based on the LM-Polygraph framework as proposed by \citet{fadeeva-etal-2023-lm}. More details are provided in Appendix \ref{app:setup}. 

\rparagraph{Baselines for UQ}
We use the following black-box UQ methods as baselines: Lexical similarity (LexSim) \citep{fomicheva-etal-2020-unsupervised}, Number of semantic sets (NumSets) \citep{lin2023generating}, Sum of eigenvalues of the graph Laplacian (EigV) \citep{lin2023generating}, Degree matrix (Deg) \citep{lin2023generating}, Eccentricity (Ecc) \citep{lin2023generating}, SelfCheckNLI (SCN) \cite{manakul-etal-2023-selfcheckgpt}. We also include three white-box methods for comparison: Maximum Sequence Probability (MSP), Monte Carlo Sequence Entropy (MCSE) \citep{malinin2020uncertainty}, and Semantic Entropy (SE) \citep{kuhn2022semantic}. More details can be found in Appendix \ref{app:baselines}. 

\subsection{Uncertainty Quantification Results}

\begin{table}[t!]
\footnotesize
\begin{tabular}{lccccc}
\toprule 
 & \textbf{FS} & \textbf{PFS} & \textbf{US} & \textbf{PUS} & \textbf{RR} \\
 \midrule
GPT-4 & 80.8 & 72.4 & 20.8 & 29.0 & 86.6 \\
GPT-3.5 & 68.3 & 68.3 & 25.7 & 25.7 & 100 \\
Yi-34B-Chat & 55.7 & 55.7 & 41.3 & 41.3 & 100 \\
Tulu-2-70B & 47.2 & 47.2 & 55.8 & 55.8 & 100 \\
Gemini 1.0 Pro & 43.2 & 42.7 & 61.7 & 62.2 & 98.9 \\
Vicuna-33B & 42.5 & 42.5 & 55.3 & 55.3 & 100 \\
\bottomrule
\end{tabular}
\caption{Results on the \textsc{FActScore-Bio}: FS and PFS are average and penalized factuality scores; US and PUS are average and penalized uncertainty scores by \textsc{Luq}; RR is the response rate. All values are percentages.}
\label{tab:factscores}
\vspace{-3mm}
\end{table}

\begin{figure*}[t!]
    \centering
    \begin{subfigure}[b]{0.25\textwidth}
        \centering
        \includegraphics[width=\textwidth]{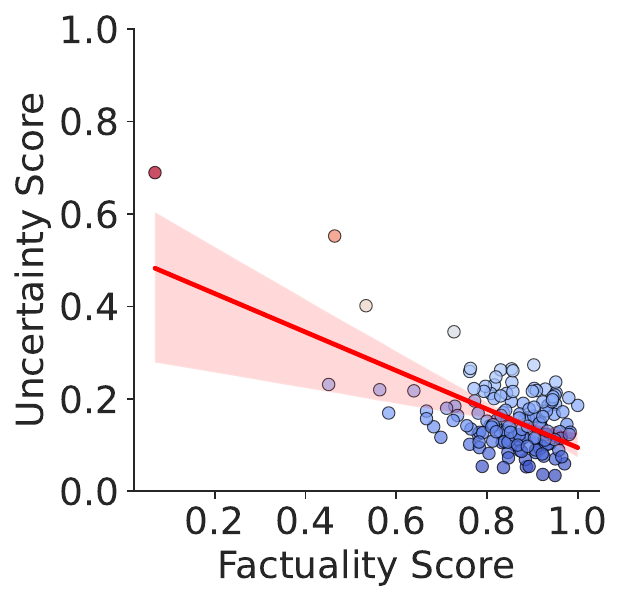}
        \caption{GPT-4}
        \label{fig:sub1}
    \end{subfigure}
    \begin{subfigure}[b]{0.25\textwidth}
        \centering
        \includegraphics[width=\textwidth]{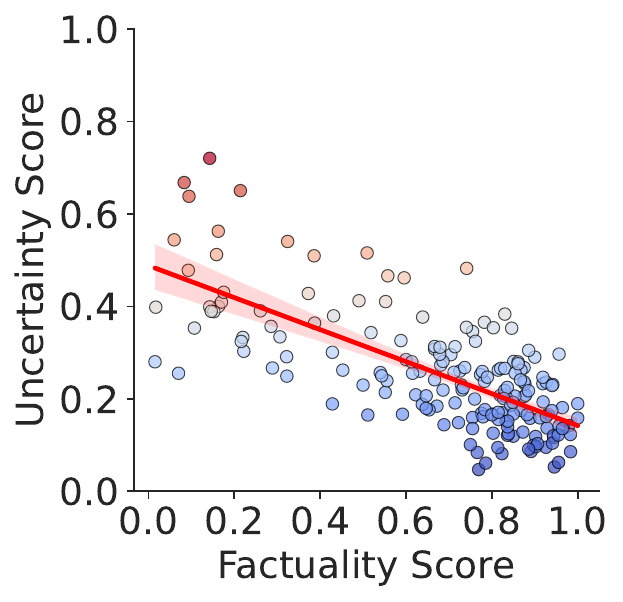}
        \caption{GPT-3.5}
        \label{fig:sub2}
    \end{subfigure}
    \begin{subfigure}[b]{0.25\textwidth}
        \centering
        \includegraphics[width=\textwidth]{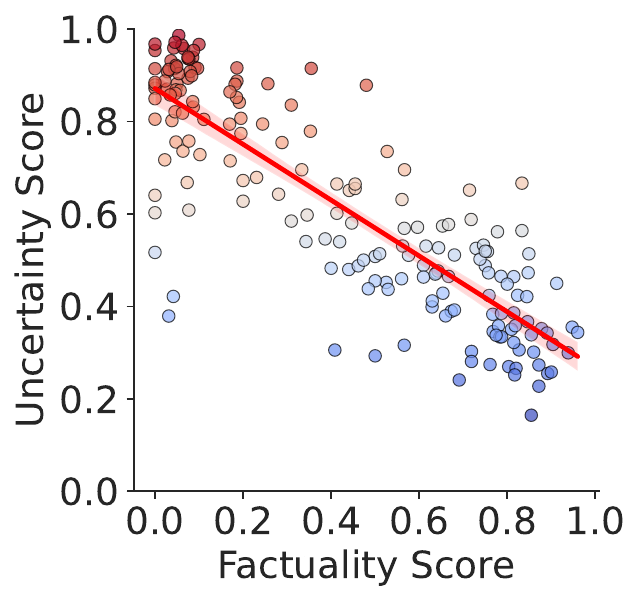}
        \caption{Gemini 1.0 Pro}
        \label{fig:sub3}
    \end{subfigure}
    \begin{subfigure}[b]{0.25\textwidth}
        \centering
        \includegraphics[width=\textwidth]{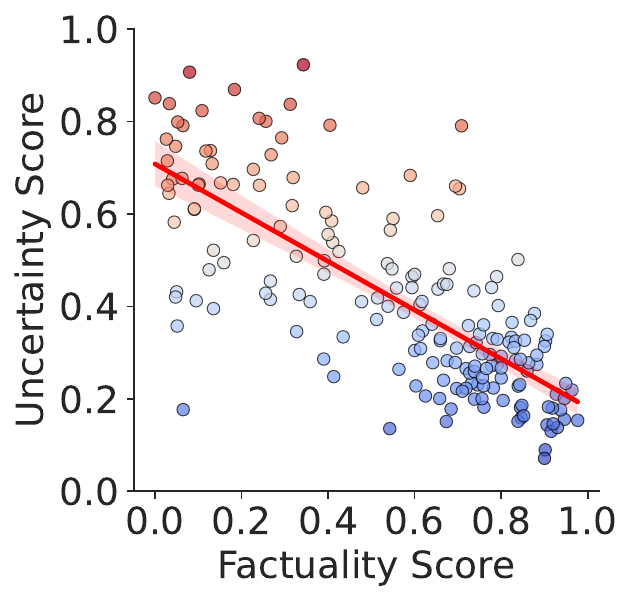}
        \caption{Yi-34B-Chat}
        \label{fig:sub4}
    \end{subfigure}
    \begin{subfigure}[b]{0.25\textwidth}
        \centering
        \includegraphics[width=\textwidth]{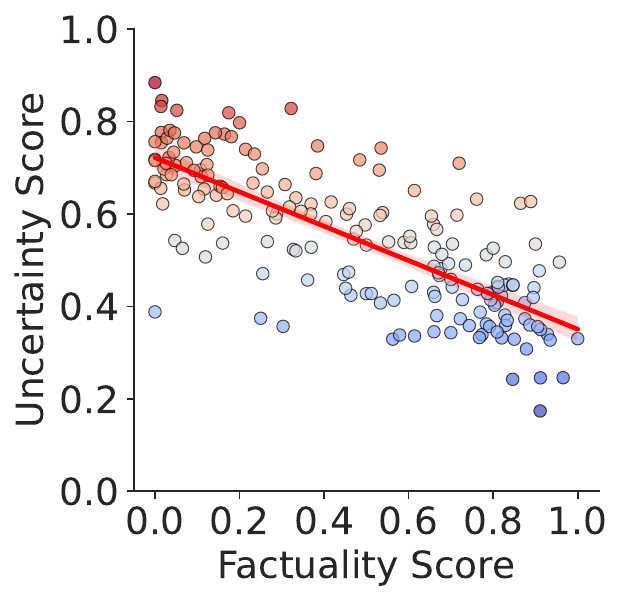}
        \caption{Tulu-2-70B}
        \label{fig:sub5}
    \end{subfigure}
    \begin{subfigure}[b]{0.25\textwidth}
        \centering
        \includegraphics[width=\textwidth]{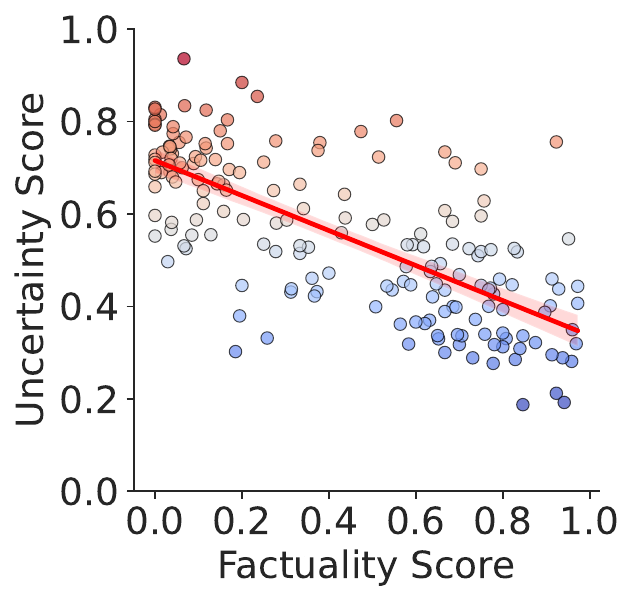}
        \caption{Vicuna-33B}
        \label{fig:sub6}
    \end{subfigure}
    \caption{Scatter plot illustrating the relationship between factuality scores (x-axis) and uncertainty scores (y-axis) for different LLMs.  Each point symbolizes an item in the FactScore dataset, with a red line highlighting the Pearson correlation. The distribution suggests a pattern where higher factuality correlates with lower uncertainty.}
    \label{fig:scatters}
    \vspace{-3mm}
\end{figure*}

\rparagraph{Effectiveness of \textsc{Luq}} 
Table \ref{tab:main_results} and Figure \ref{fig:scatters} illustrate the correlation between factuality scores and uncertainty scores. The results highlight \textsc{Luq}'s effectiveness as an indicator of model factuality in long text generation tasks. \textsc{Luq} demonstrates a strong negative correlation for GPT-3.5, Gemini 1.0 Pro, Yi-34B-Chat, Vicuna-33B, and Tulu-2-70B, with the strongest Pearson correlation being -0.851.
For the baseline methods, LexSim emerges as a robust baseline offering lower computational demands. 
The confidence-based SCN method demonstrates the best Spearman correlation in models such as Gemini 1.0 Pro and Tulu-2-70B. Other baselines such as Ecc, NumSets and Deg yield unsatisfactory results, occasionally exhibiting even positive correlations. Case studies of our proposed \textsc{Luq} method can be found in Appendix \ref{app:case_study}.

\rparagraph{Variations of \textsc{Luq}} We compare the original \textsc{Luq} with its two variations, \textsc{Luq-Atomic} and \textsc{Luq-Pair}, in Table \ref{tab:luq_variations}. We find that, with more fine-grained entailment checking, both consistently outperform the original \textsc{Luq}. Further discussion on the pros and cons of these variations, along with usage guidelines, can be found in Appendix \ref{app:variations}.

\rparagraph{\textsc{Luq} for GPT-4} We also observe that \textsc{Luq} is better suited for models with relatively lower factuality and a lack of self-expressiveness regarding uncertainty. For models with high factuality capabilities, such as GPT-4, \textsc{Luq} only demonstrates a moderate correlation with factuality scores. As shown in Table \ref{tab:factscores}, among all models, GPT-4 exhibits the highest overall factuality scores and the lowest average uncertainty scores. Figure \ref{fig:sub1} also shows that the data points of GPT-4 are tightly clustered with only few instances of high uncertainty. This is because GPT-4 tends to abstain from answering questions more often compared to other models, highlighting improved uncertainty self-detection. However, this observation does not influence the effectiveness of our method, as in real life models with lower factuality and unable to express uncertainty are in greater need of external uncertainty measurements. 

\rparagraph{\textsc{Luq} in \textsc{FActScore-Dis}} We test one closed-source LLM, GPT-3.5, and one open-source LLM, Yi-34B-Chat in our newly proposed \textsc{FActScore-Dis}. Our \textsc{Luq} model consistently surpasses the performance of baseline models, thereby demonstrating its effectiveness on the newly proposed dataset within the medical domain. 

\begin{figure}[ht!]
    \centering
    \includegraphics[width=\columnwidth]{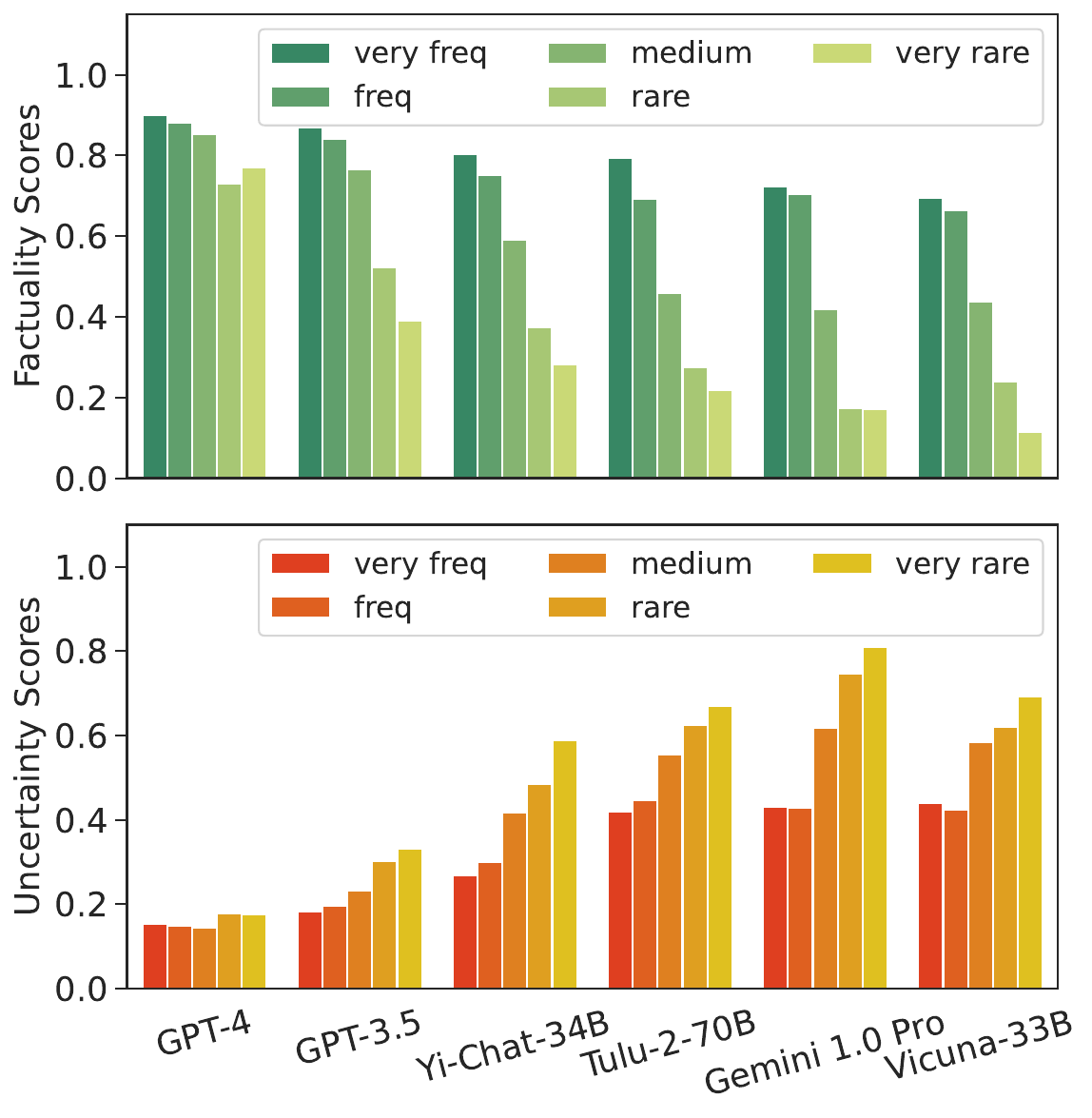}
    \vspace{-4mm}
    \caption{Factuality and uncertainty scores across different frequencies on \textsc{FActScore-Bio}. 
    }
    \label{fig:frequency}
    \vspace{-3mm}
\end{figure}

\rparagraph{Higher frequency leads to higher factuality and lower uncertainty} 
In Figure \ref{fig:frequency}, we compare the factuality and uncertainty scores across different entity frequencies. The original \textsc{FActScore} dataset provides the frequency of each entity in Wikipedia, categorizing them based on page views and co-occurrence within the training set \citep{min-etal-2023-factscore}. Frequencies are classified into five categories, ranging from ``very rare'' to ``very frequent.'' Our observations suggest that questions associated with higher entity frequencies tend to yield more factual responses, alongside decreased model uncertainty. Notably, GPT-4 demonstrates consistent performance regarding uncertainty and factuality across varying frequencies, potentially attributable to its selective response strategy. Although it answers all the questions in the ``very frequent,'' ``frequent,'' and ``medium'' categories, it refuses to answer around 25\% of ``rare'' questions and 30\% of ``very rare'' questions.

\begin{table*}[ht!]
\centering
\footnotesize	
\begin{tabular}{ccccccccccc}
\toprule
 & \multicolumn{2}{c}{GPT-3.5} & \multicolumn{2}{c}{Yi-34B-Chat} & \multicolumn{2}{c}{Tulu-2-70B} & \multicolumn{2}{c}{Vicuna-33B} & \multicolumn{2}{c}{Gemini 1.0 Pro} \\

\cmidrule(lr){2-3} \cmidrule(lr){4-5} \cmidrule(lr){6-7} \cmidrule(lr){8-9} \cmidrule(lr){10-11}

\multirow{-2}{*}{\textbf{Percentile}} & FS & US & FS & US & FS & US & FS & US & FS & US \\

\hline
\textbf{0} & \cellcolor[HTML]{EFEFEF}68.3 & 25.7 & \cellcolor[HTML]{EFEFEF}55.7 & 41.3 & \cellcolor[HTML]{EFEFEF}47.2 & 55.8 & \cellcolor[HTML]{EFEFEF}42.5 & 55.3 & \cellcolor[HTML]{EFEFEF}43.2 & 61.7 \\
\textbf{2.5} & \cellcolor[HTML]{EFEFEF}69.8 & 24.1 & \cellcolor[HTML]{EFEFEF}56.9 & 40.2 & \cellcolor[HTML]{EFEFEF}48.3 & 53.9 & \cellcolor[HTML]{EFEFEF}43.6 & 54.4 & \cellcolor[HTML]{EFEFEF}44.3 & 60.0 \\
\textbf{5} & \cellcolor[HTML]{EFEFEF}70.8 & 23.4 & \cellcolor[HTML]{EFEFEF}58.0 & 39.3 & \cellcolor[HTML]{EFEFEF}49.4 & 53.1 & \cellcolor[HTML]{EFEFEF}44.5 & 53.8 & \cellcolor[HTML]{EFEFEF}45.2 & 59.2 \\
\textbf{7.5} & \cellcolor[HTML]{EFEFEF}71.5 & 22.7 & \cellcolor[HTML]{EFEFEF}58.9 & 38.1 & \cellcolor[HTML]{EFEFEF}50.3 & 52.6 & \cellcolor[HTML]{EFEFEF}45.5 & 53.0 & \cellcolor[HTML]{EFEFEF}46.3 & 58.2 \\
\textbf{10} & \cellcolor[HTML]{EFEFEF}72.3 & 22.2 & \cellcolor[HTML]{EFEFEF}60.2 & 36.8 & \cellcolor[HTML]{EFEFEF}51.4 & 51.9 & \cellcolor[HTML]{EFEFEF}46.1 & 52.3 & \cellcolor[HTML]{EFEFEF}47.3 & 57.7 \\
\textbf{12.5} & \cellcolor[HTML]{EFEFEF}74.1 & 21.6 & \cellcolor[HTML]{EFEFEF}61.7 & 35.0 & \cellcolor[HTML]{EFEFEF}52.1 & 51.3 & \cellcolor[HTML]{EFEFEF}46.5 & 51.6 & \cellcolor[HTML]{EFEFEF}48.4 & 56.3 \\
\textbf{15} & \cellcolor[HTML]{EFEFEF}75.0 & 21.2 & \cellcolor[HTML]{EFEFEF}62.9 & 34.2 & \cellcolor[HTML]{EFEFEF}53.3 & 50.6 & \cellcolor[HTML]{EFEFEF}47.5 & 51.0 & \cellcolor[HTML]{EFEFEF}49.5 & 55.4 \\
\bottomrule
\end{tabular}
\caption{Selective question answering results on \textsc{FActScore-Bio} (expressed as percentage). The percentile indicates the percentage of questions for which answers were abstained. }
\label{tab:selective}
\end{table*}

\begin{table}[ht!]
\centering
\footnotesize
\tabcolsep=2.5mm
\begin{tabular}{lccc}
\toprule
\textbf{Methods} & PFS & PUS & AD \\

\midrule

Tulu-2-70B & 47.2 & 55.8 & 42.1 \\
Gemini 1.0 Pro & 42.7 & 62.2 & 29.5 \\
Vicuna-33B & 42.5 & 58.1 & 28.4 \\
\rowcolor{LightCyan}
\textsc{Luq-Ensemble} & \textbf{52.8} & \textbf{45.8} & 100 \\
\midrule \midrule

Yi-34B-Chat & 55.7 & 41.3 & 66.1 \\
Tulu-2-70B & 47.2 & 55.8 & 21.3 \\
Gemini 1.0 Pro & 42.7 & 62.2 & 12.6 \\
\rowcolor{LightCyan}
\textsc{Luq-Ensemble} & \textbf{58.8} & \textbf{37.6} & 100 \\
\midrule \midrule

GPT-3.5 & 67.3 & 25.7 & 92.4 \\
Gemini 1.0 Pro & 42.7 & 62.2 & 1.64 \\
Vicuna-33B & 42.5 & 58.1 & 6.01 \\
\rowcolor{LightCyan}
\textsc{Luq-Ensemble} & \textbf{67.4} & \textbf{24.8} & 100 \\
\midrule \midrule

GPT-4 & 72.1 & 29.0 & 60.1 \\
GPT-3.5 & 67.3 & 25.7 & 32.8 \\
Yi-34B-Chat & 55.7 & 41.3 & 7.10 \\
\rowcolor{LightCyan}
\textsc{Luq-Ensemble} & \textbf{76.6} & \textbf{17.3} & 100 \\
\bottomrule
\end{tabular}
\caption{Results of different ensemble strategies on \textsc{FActScore-Bio} (expressed as percentage). The Answer Distribution (AD) indicates the percentage of final answers generated by each component model. }
\label{tab:ensemble}
\end{table}

\subsection{\textsc{Luq-Ensemble}} 

Given the variance in training corpus, different LLMs may possess varying levels of knowledge for a specific question. 
After obtaining outputs from different LLMs, the challenge now is to \textit{choose the best one without fact-checking each answer} (which is both time-consuming and costly \citep{guo-etal-2022-survey, zhang2024need}). Utilizing the \textsc{Luq} uncertainty score as a reliable indicator of factuality, we enhance overall performance through an ensemble approach. In this method, the model exhibiting the lowest \textsc{Luq} score for a given question is chosen as the final answer. Experimental results (Table \ref{tab:ensemble}) affirm the superiority of the \textsc{Luq-Ensemble} over its constituent counterparts. 

\rparagraph{Ensembling models with similar factuality scores can notably enhance performance} Our findings suggest that ensembling models with similar factuality scores can significantly enhance performance. For instance, in the combination of Tulu-2-70B, Gemini 1.0 Pro, and Vicuna-33B, the PFS increases by 5\% compared to the originally top-performing Tulu-2-70B, which scored 47.19\%. Additionally, ensembling models with comparable performance leads to a more balanced distribution of answers. In contrast, integrating a model with substantially superior performance, as seen in the combination of GPT-3.5, Gemini 1.0 Pro, and Vicuna-33B, predominantly favors answers from GPT-3.5 (92.35\%), leading to marginal improvement (0.06\%) in the ensemble method.


\rparagraph{Ensembling does not guarantee better performance} While ensembling always reduces uncertainty scores (as we select the model with the least uncertainty), it does not necessarily improve factuality scores. Ensembling LLMs according to poor UQ methods may result in overall performance that is worse than that of its individual components. Table \ref{tab:different_ensembles} in Appendix \ref{app:ensemble} compares the effectiveness of using LUQ as the ensemble indicator with other methods. \textit{The results indicate that ensembling does not inherently enhance performance.} For example, with UQ method Ecc, the ensemble factuality score can be lower than that of its best-performing component (47.2\% vs 43.3\%). In contrast, using \textsc{Luq} as the ensembling indicator yields the best overall performance.

\subsection{Selective Question Answering}

From Table \ref{tab:factscores}, it is observed that while GPT-4 opts not to respond to some queries, other models generally attempt to answer all questions. The limited refusal by Gemini 1.0 Pro primarily stems from considerations of sensitive content and regulatory constraints, rather than uncertainty.
Therefore, we investigate the application of the \textsc{Luq} score to equip these models with the capability for selective question answering—that is, to enable them to decline responses when uncertain. Contrary to the traditional aim of responding correctly to every question, the objective in a selective question answering framework is to preserve accuracy while maximizing the number of questions answered \citep{kamath-etal-2020-selective, cole-etal-2023-selectively, yang2023uncertaintyaware, dong2024llmpersonalizedjudge}.

Table \ref{tab:selective} presents the results of selective question answering. The models are permitted to refrain from answering questions with high uncertainty. The \textit{percentiles} indicate the proportion of questions each model abstained from answering. The findings demonstrate that adopting a selective answering approach enhances the models' factuality by allowing for more question rejections. By declining to answer a similar proportion of questions (approximately 15\%) as GPT-4, the models typically achieve an improvement of over 5\% in overall factuality scores. In Appendix \ref{app:selective_qa}, we provide a detailed discussion on how practitioners can use \textsc{Luq} to implement selective answering strategies, including setting and adjusting uncertainty thresholds.

\section{Related Work}

\rparagraph{UQ in Machine Learning} Prior to LLMs, UQ has been extensively explored within the field of machine learning \citep{gawlikowski2023survey}. According to the source of uncertainty, it is typically categorized into two types: aleatoric and epistemic uncertainty\citep{hora1996aleatory, der2009aleatory}. Aleatoric uncertainty, also known as statistical uncertainty, pertains to the inherent randomness in experimental outcomes due to stochastic effects \citep{hullermeier2021aleatoric}. In contrast, epistemic uncertainty stems from incomplete knowledge, potentially including uncertainties in a machine learning model's parameters or the lack of certain training data \citep{hullermeier2021aleatoric, huang2023look}. Our focus is primarily on epistemic uncertainty.


\rparagraph{UQ in LLMs} In contrast to discriminative models, which readily provide probability scores for specific categories, uncertainty estimation in generative LLMs presents unique challenges: (1) There is an exponential increase in the output space as sentence length grows, rendering the evaluation of all possible predictions impractical \citep{geng2023survey, wang2023survey}. (2) The significance of semantic nuances and their inherent uncertainties, which diverges from the fixed category labels typical of discriminative models, complicates matters further \citep{kuhn2022semantic}. Generally, UQ methods for LLMs can be categorized based on the accessibility of the model's internal states, distinguishing between black-box and white-box approaches. White-box LLMs often rely on logit-based evaluations, assessing sentence uncertainty through token-level probabilities or entropy \citep{murray-chiang-2018-correcting, kuhn2022semantic, vazhentsev-etal-2023-efficient, duan2023shifting}.

However, as access to LLMs increasingly relies on API calls, research has pivoted towards black-box methods. These can be further categorized into: (i) \textit{verbalized methods}, which prompt LLMs to articulate their uncertainty in the output, using phrases like ``I am sure" or ``I do not know" \citep{mielke-etal-2022-reducing}. Nonetheless, a practical mismatch between the expressed and actual uncertainty levels has been noted \citep{lin2022teaching, xiong2023llms}. \citet{xiong2023llms} highlight that LLMs often display excessive confidence when verbalizing their certainty. (ii) \textit{Consistency-based (sampling-based)} estimation premises on the assumption that increased uncertainty in a model corresponds to greater diversity in its outputs, frequently resulting in hallucinatory outputs \citep{manakul-etal-2023-selfcheckgpt, lin2023generating}. Our proposed method, \textsc{Luq}, follows this consistency-based approach. There are also efforts on integrating verbalized methods with consistency-based approaches \citep{xiong2023llms, rivera2024combining}. Understanding uncertainty in LLMs can enhance in-context learning \citep{zhou2023batch, li2023task}, selective question answering~\citep{yang2023uncertaintyaware}, LLM cascading \citep{huang2024calibrating}, adaptive retrieval~\citep{ding2024retrieve}, language agents~\citep{han-etal-2024-towards}, and model self-refinement~\citep{yao2024learning, chen2024self}.
\section{Conclusion}

In this work, we first identify that existing UQ methods are ineffective on long text generation. We therefore introduce \textsc{Luq}, a novel UQ method tailored for long-form text generation in LLMs. It overcomes the limitation of previous methods by calculating sentence level consistency. We conduct extensive experiments over six popular LLMs, such as GPT-4 and Gemini 1.0 Pro. We extend the existing \textsc{FActScore} dataset with human validation and annotations for additional disease domain. Our findings demonstrate that \textsc{Luq} significantly improves the correlation with models' factuality scores over previous methods across various different setups and domains. \textsc{Luq} serves as a reliable indicator of model's factuality performance. Additionally, we present \textsc{Luq-Ensemble}, a model ensembling and selective question answering strategy, which showcases a promising avenue for enhancing the factual accuracy of LLM outputs. This research not only advances our understanding of UQ in the context of LLMs but also offers practical tools for improving the reliability and trustworthiness of AI-generated content.

\section*{Limitation}

The limitations of this study include the following: \textbf{(1)} A primary challenge in studying uncertainty quantification for long text generation lies in the difficulty of evaluating the generated text. Unlike classification tasks and short-answer QA, there is no straightforward metric for assessing the quality of generated text. In this study, we employ the factuality score as the primary evaluation metric, thereby leaving other text aspects, such as coherence, cohesion, and creativity, under-explored. Future work could investigate uncertainty scores using more comprehensive evaluation metrics. \textbf{(2)} In this study, we do not investigate the performance of UQ methods under ambiguous and unanswerable questions, such as ASQA \citep{stelmakh-etal-2022-asqa} and SelfAware \citep{yin-etal-2023-large}. Previous uncertainty metrics for short-answer questions are tested on answerable questions with clear intentions. This is because clearly defined questions with definite answers provide a straightforward framework for evaluating model accuracy. In contrast, unanswerable or ambiguous questions lack clear ground truths, complicating the assessment of uncertainty estimates. We advocate researchers to explore more in this area in the future. \textbf{(3)} We focus on assessing the overall uncertainty of a model, rather than model uncertainty on individual instances. The relative uncertainty equation in Section 2.2 represents an ideal scenario. If a model learns a significant amount of non-factual data over factual data for a particular entity/instance, the aforementioned equation can be inaccurate for that case. Future work could investigate the causes of this special case and develop strategies to address it during the pre-training stage.

\section*{Ethics Statement}

Our research adheres to rigorous ethical guidelines, with a strong emphasis on data privacy, bias mitigation, and societal impact. During the dataset construction phase, we verified the licenses of all software utilized and ensured strict compliance with these licenses. We have thoroughly assessed our project and do not foresee any other potential risks.

\section*{Acknowledgments}

Caiqi Zhang is supported by an Amazon Studentship. We thank Chengzu Li, Yulong Chen, and the reviewers for their valuable feedback on this paper.


\bibliography{anthology,custom}

\begin{thebibliography}{59}
\expandafter\ifx\csname natexlab\endcsname\relax\def\natexlab#1{#1}\fi

\bibitem[{{01.ai}(2023)}]{yi34bchat}
{01.ai}. 2023.
\newblock Building the next generation of open-source and bilingual llms.
\newblock \url{https://huggingface.co/01-ai/Yi-34B-Chat}.
\newblock Accessed: 2024-02-05.

\bibitem[{Anthropic(2023)}]{anthropic2023claude}
Anthropic. 2023.
\newblock Introducing claude 2.1.
\newblock Available from Anthropic: \url{https://www.anthropic.com/news/claude-2-1}.

\bibitem[{Baan et~al.(2023)Baan, Daheim, Ilia, Ulmer, Li, Fernández, Plank, Sennrich, Zerva, and Aziz}]{baan2023uncertainty}
Joris Baan, Nico Daheim, Evgenia Ilia, Dennis Ulmer, Haau-Sing Li, Raquel Fernández, Barbara Plank, Rico Sennrich, Chrysoula Zerva, and Wilker Aziz. 2023.
\newblock \href {http://arxiv.org/abs/2307.15703} {Uncertainty in natural language generation: From theory to applications}.

\bibitem[{Chang et~al.(2023)Chang, Wang, Wang, Wu, Yang, Zhu, Chen, Yi, Wang, Wang et~al.}]{chang2023survey}
Yupeng Chang, Xu~Wang, Jindong Wang, Yuan Wu, Linyi Yang, Kaijie Zhu, Hao Chen, Xiaoyuan Yi, Cunxiang Wang, Yidong Wang, et~al. 2023.
\newblock \href {https://arxiv.org/abs/2307.03109} {A survey on evaluation of large language models}.
\newblock \emph{ArXiv preprint}, abs/2307.03109.

\bibitem[{Chen and Mueller(2023)}]{chen2023quantifying}
Jiuhai Chen and Jonas Mueller. 2023.
\newblock \href {http://arxiv.org/abs/2308.16175} {Quantifying uncertainty in answers from any language model and enhancing their trustworthiness}.

\bibitem[{Chen et~al.(2024)Chen, Liu, Yan, Bai, Zhong, Yang, Yang, Zhu, and Zhang}]{chen2024self}
Yulong Chen, Yang Liu, Jianhao Yan, Xuefeng Bai, Ming Zhong, Yinghao Yang, Ziyi Yang, Chenguang Zhu, and Yue Zhang. 2024.
\newblock \href {https://arxiv.org/abs/2408.08978} {See what llms cannot answer: A self-challenge framework for uncovering llm weaknesses}.

\bibitem[{Cole et~al.(2023)Cole, Zhang, Gillick, Eisenschlos, Dhingra, and Eisenstein}]{cole-etal-2023-selectively}
Jeremy Cole, Michael Zhang, Daniel Gillick, Julian Eisenschlos, Bhuwan Dhingra, and Jacob Eisenstein. 2023.
\newblock \href {https://doi.org/10.18653/v1/2023.emnlp-main.35} {Selectively answering ambiguous questions}.
\newblock In \emph{Proceedings of the 2023 Conference on Empirical Methods in Natural Language Processing}, pages 530--543, Singapore. Association for Computational Linguistics.

\bibitem[{Der~Kiureghian and Ditlevsen(2009)}]{der2009aleatory}
Armen Der~Kiureghian and Ove Ditlevsen. 2009.
\newblock \href {https://www.sciencedirect.com/science/article/pii/S0167473008000556} {Aleatory or epistemic? does it matter?}
\newblock \emph{Structural safety}, 31(2):105--112.

\bibitem[{Ding et~al.(2024)Ding, Pang, Wei, Shen, and Cheng}]{ding2024retrieve}
Hanxing Ding, Liang Pang, Zihao Wei, Huawei Shen, and Xueqi Cheng. 2024.
\newblock \href {http://arxiv.org/abs/2402.10612} {Retrieve only when it needs: Adaptive retrieval augmentation for hallucination mitigation in large language models}.

\bibitem[{Dong et~al.(2024)Dong, Hu, and Collier}]{dong2024llmpersonalizedjudge}
Yijiang~River Dong, Tiancheng Hu, and Nigel Collier. 2024.
\newblock \href {https://arxiv.org/abs/2406.11657} {Can llm be a personalized judge?}

\bibitem[{Duan et~al.(2023)Duan, Cheng, Wang, Wang, Zavalny, Xu, Kailkhura, and Xu}]{duan2023shifting}
Jinhao Duan, Hao Cheng, Shiqi Wang, Chenan Wang, Alex Zavalny, Renjing Xu, Bhavya Kailkhura, and Kaidi Xu. 2023.
\newblock \href {https://arxiv.org/abs/2307.01379} {Shifting attention to relevance: Towards the uncertainty estimation of large language models}.
\newblock \emph{ArXiv preprint}, abs/2307.01379.

\bibitem[{Fadeeva et~al.(2023)Fadeeva, Vashurin, Tsvigun, Vazhentsev, Petrakov, Fedyanin, Vasilev, Goncharova, Panchenko, Panov, Baldwin, and Shelmanov}]{fadeeva-etal-2023-lm}
Ekaterina Fadeeva, Roman Vashurin, Akim Tsvigun, Artem Vazhentsev, Sergey Petrakov, Kirill Fedyanin, Daniil Vasilev, Elizaveta Goncharova, Alexander Panchenko, Maxim Panov, Timothy Baldwin, and Artem Shelmanov. 2023.
\newblock \href {https://doi.org/10.18653/v1/2023.emnlp-demo.41} {{LM}-polygraph: Uncertainty estimation for language models}.
\newblock In \emph{Proceedings of the 2023 Conference on Empirical Methods in Natural Language Processing: System Demonstrations}, pages 446--461, Singapore. Association for Computational Linguistics.

\bibitem[{Fan et~al.(2019)Fan, Jernite, Perez, Grangier, Weston, and Auli}]{fan-etal-2019-eli5}
Angela Fan, Yacine Jernite, Ethan Perez, David Grangier, Jason Weston, and Michael Auli. 2019.
\newblock \href {https://doi.org/10.18653/v1/P19-1346} {{ELI}5: Long form question answering}.
\newblock In \emph{Proceedings of the 57th Annual Meeting of the Association for Computational Linguistics}, pages 3558--3567, Florence, Italy. Association for Computational Linguistics.

\bibitem[{Fomicheva et~al.(2020)Fomicheva, Sun, Yankovskaya, Blain, Guzm{\'a}n, Fishel, Aletras, Chaudhary, and Specia}]{fomicheva-etal-2020-unsupervised}
Marina Fomicheva, Shuo Sun, Lisa Yankovskaya, Fr{\'e}d{\'e}ric Blain, Francisco Guzm{\'a}n, Mark Fishel, Nikolaos Aletras, Vishrav Chaudhary, and Lucia Specia. 2020.
\newblock \href {https://doi.org/10.1162/tacl_a_00330} {Unsupervised quality estimation for neural machine translation}.
\newblock \emph{Transactions of the Association for Computational Linguistics}, 8:539--555.

\bibitem[{Gawlikowski et~al.(2023)Gawlikowski, Tassi, Ali, Lee, Humt, Feng, Kruspe, Triebel, Jung, Roscher et~al.}]{gawlikowski2023survey}
Jakob Gawlikowski, Cedrique Rovile~Njieutcheu Tassi, Mohsin Ali, Jongseok Lee, Matthias Humt, Jianxiang Feng, Anna Kruspe, Rudolph Triebel, Peter Jung, Ribana Roscher, et~al. 2023.
\newblock \href {https://link.springer.com/article/10.1007/s10462-023-10562-9} {A survey of uncertainty in deep neural networks}.
\newblock \emph{Artificial Intelligence Review}, 56(Suppl 1):1513--1589.

\bibitem[{Geifman and El{-}Yaniv(2017)}]{geifman2017selective}
Yonatan Geifman and Ran El{-}Yaniv. 2017.
\newblock \href {https://proceedings.neurips.cc/paper/2017/hash/4a8423d5e91fda00bb7e46540e2b0cf1-Abstract.html} {Selective classification for deep neural networks}.
\newblock In \emph{Advances in Neural Information Processing Systems 30: Annual Conference on Neural Information Processing Systems 2017, December 4-9, 2017, Long Beach, CA, {USA}}, pages 4878--4887.

\bibitem[{{Gemini Team}(2023)}]{Gemini2023}
{Gemini Team}. 2023.
\newblock \href {https://storage.googleapis.com/deepmind-media/gemini/gemini_1_report.pdf} {Gemini: A family of highly capable multimodal models}.
\newblock Technical report, Google.

\bibitem[{Geng et~al.(2023)Geng, Cai, Wang, Koeppl, Nakov, and Gurevych}]{geng2023survey}
Jiahui Geng, Fengyu Cai, Yuxia Wang, Heinz Koeppl, Preslav Nakov, and Iryna Gurevych. 2023.
\newblock \href {https://arxiv.org/abs/2311.08298} {A survey of language model confidence estimation and calibration}.
\newblock \emph{ArXiv preprint}, abs/2311.08298.

\bibitem[{Guo et~al.(2022)Guo, Schlichtkrull, and Vlachos}]{guo-etal-2022-survey}
Zhijiang Guo, Michael Schlichtkrull, and Andreas Vlachos. 2022.
\newblock \href {https://doi.org/10.1162/tacl_a_00454} {A survey on automated fact-checking}.
\newblock \emph{Transactions of the Association for Computational Linguistics}, 10:178--206.

\bibitem[{Han et~al.(2024)Han, Buntine, and Shareghi}]{han-etal-2024-towards}
Jiuzhou Han, Wray Buntine, and Ehsan Shareghi. 2024.
\newblock \href {https://doi.org/10.18653/v1/2024.findings-acl.398} {Towards uncertainty-aware language agent}.
\newblock In \emph{Findings of the Association for Computational Linguistics ACL 2024}, pages 6662--6685, Bangkok, Thailand and virtual meeting. Association for Computational Linguistics.

\bibitem[{He et~al.(2023)He, Gao, and Chen}]{he2022debertav3}
Pengcheng He, Jianfeng Gao, and Weizhu Chen. 2023.
\newblock \href {https://openreview.net/pdf?id=sE7-XhLxHA} {Debertav3: Improving deberta using electra-style pre-training with gradient-disentangled embedding sharing}.
\newblock In \emph{The Eleventh International Conference on Learning Representations, {ICLR} 2023, Kigali, Rwanda, May 1-5, 2023}. OpenReview.net.

\bibitem[{Hendrycks and Gimpel(2017)}]{hendrycks2016baseline}
Dan Hendrycks and Kevin Gimpel. 2017.
\newblock \href {https://openreview.net/forum?id=Hkg4TI9xl} {A baseline for detecting misclassified and out-of-distribution examples in neural networks}.
\newblock In \emph{5th International Conference on Learning Representations, {ICLR} 2017, Toulon, France, April 24-26, 2017, Conference Track Proceedings}. OpenReview.net.

\bibitem[{Hora(1996)}]{hora1996aleatory}
Stephen~C Hora. 1996.
\newblock \href {https://www.sciencedirect.com/science/article/pii/S0951832096000774} {Aleatory and epistemic uncertainty in probability elicitation with an example from hazardous waste management}.
\newblock \emph{Reliability Engineering \& System Safety}, 54(2-3):217--223.

\bibitem[{Hu et~al.(2023)Hu, Zhang, Zhao, Huang, and Wu}]{hu2023uncertainty}
Mengting Hu, Zhen Zhang, Shiwan Zhao, Minlie Huang, and Bingzhe Wu. 2023.
\newblock \href {http://arxiv.org/abs/2306.04459} {Uncertainty in natural language processing: Sources, quantification, and applications}.

\bibitem[{Huang et~al.(2024{\natexlab{a}})Huang, Li, Yu, Sesia, Hassani, Lee, Bastani, and Dobriban}]{huang2024uncertainty}
Xinmeng Huang, Shuo Li, Mengxin Yu, Matteo Sesia, Hamed Hassani, Insup Lee, Osbert Bastani, and Edgar Dobriban. 2024{\natexlab{a}}.
\newblock \href {http://arxiv.org/abs/2404.03163} {Uncertainty in language models: Assessment through rank-calibration}.

\bibitem[{Huang et~al.(2023)Huang, Song, Wang, Chen, and Ma}]{huang2023look}
Yuheng Huang, Jiayang Song, Zhijie Wang, Huaming Chen, and Lei Ma. 2023.
\newblock \href {https://arxiv.org/abs/2307.10236} {Look before you leap: An exploratory study of uncertainty measurement for large language models}.
\newblock \emph{ArXiv preprint}, abs/2307.10236.

\bibitem[{Huang et~al.(2024{\natexlab{b}})Huang, Liu, Thirukovalluru, Cohan, and Dhingra}]{huang2024calibrating}
Yukun Huang, Yixin Liu, Raghuveer Thirukovalluru, Arman Cohan, and Bhuwan Dhingra. 2024{\natexlab{b}}.
\newblock \href {http://arxiv.org/abs/2402.06544} {Calibrating long-form generations from large language models}.

\bibitem[{H{\"u}llermeier and Waegeman(2021)}]{hullermeier2021aleatoric}
Eyke H{\"u}llermeier and Willem Waegeman. 2021.
\newblock \href {https://link.springer.com/article/10.1007/s10994-021-05946-3} {Aleatoric and epistemic uncertainty in machine learning: An introduction to concepts and methods}.
\newblock \emph{Machine Learning}, 110:457--506.

\bibitem[{Ivison et~al.(2023)Ivison, Wang, Pyatkin, Lambert, Peters, Dasigi, Jang, Wadden, Smith, Beltagy, and Hajishirzi}]{ivison2023camels}
Hamish Ivison, Yizhong Wang, Valentina Pyatkin, Nathan Lambert, Matthew Peters, Pradeep Dasigi, Joel Jang, David Wadden, Noah~A. Smith, Iz~Beltagy, and Hannaneh Hajishirzi. 2023.
\newblock \href {http://arxiv.org/abs/2311.10702} {Camels in a changing climate: Enhancing lm adaptation with tulu 2}.

\bibitem[{Kamath et~al.(2020)Kamath, Jia, and Liang}]{kamath-etal-2020-selective}
Amita Kamath, Robin Jia, and Percy Liang. 2020.
\newblock \href {https://doi.org/10.18653/v1/2020.acl-main.503} {Selective question answering under domain shift}.
\newblock In \emph{Proceedings of the 58th Annual Meeting of the Association for Computational Linguistics}, pages 5684--5696, Online. Association for Computational Linguistics.

\bibitem[{Kuhn et~al.(2023)Kuhn, Gal, and Farquhar}]{kuhn2022semantic}
Lorenz Kuhn, Yarin Gal, and Sebastian Farquhar. 2023.
\newblock \href {https://openreview.net/pdf?id=VD-AYtP0dve} {Semantic uncertainty: Linguistic invariances for uncertainty estimation in natural language generation}.
\newblock In \emph{The Eleventh International Conference on Learning Representations, {ICLR} 2023, Kigali, Rwanda, May 1-5, 2023}. OpenReview.net.

\bibitem[{Landis and Koch(1977)}]{landis1977measurement}
J~Richard Landis and Gary~G Koch. 1977.
\newblock The measurement of observer agreement for categorical data.
\newblock \emph{biometrics}, pages 159--174.

\bibitem[{Li et~al.(2023)Li, Zhou, Glavaš, Korhonen, and Vulić}]{li2023task}
Chengzu Li, Han Zhou, Goran Glavaš, Anna Korhonen, and Ivan Vulić. 2023.
\newblock \href {http://arxiv.org/abs/2312.13772} {On task performance and model calibration with supervised and self-ensembled in-context learning}.

\bibitem[{Lin et~al.(2022)Lin, Hilton, and Evans}]{lin2022teaching}
Stephanie Lin, Jacob Hilton, and Owain Evans. 2022.
\newblock \href {https://arxiv.org/abs/2205.14334} {Teaching models to express their uncertainty in words}.
\newblock \emph{ArXiv preprint}, abs/2205.14334.

\bibitem[{Lin et~al.(2023)Lin, Trivedi, and Sun}]{lin2023generating}
Zhen Lin, Shubhendu Trivedi, and Jimeng Sun. 2023.
\newblock \href {http://arxiv.org/abs/2305.19187} {Generating with confidence: Uncertainty quantification for black-box large language models}.

\bibitem[{Malinin and Gales(2021)}]{malinin2020uncertainty}
Andrey Malinin and Mark J.~F. Gales. 2021.
\newblock \href {https://openreview.net/forum?id=jN5y-zb5Q7m} {Uncertainty estimation in autoregressive structured prediction}.
\newblock In \emph{9th International Conference on Learning Representations, {ICLR} 2021, Virtual Event, Austria, May 3-7, 2021}. OpenReview.net.

\bibitem[{Manakul et~al.(2023)Manakul, Liusie, and Gales}]{manakul-etal-2023-selfcheckgpt}
Potsawee Manakul, Adian Liusie, and Mark Gales. 2023.
\newblock \href {https://doi.org/10.18653/v1/2023.emnlp-main.557} {{S}elf{C}heck{GPT}: Zero-resource black-box hallucination detection for generative large language models}.
\newblock In \emph{Proceedings of the 2023 Conference on Empirical Methods in Natural Language Processing}, pages 9004--9017, Singapore. Association for Computational Linguistics.

\bibitem[{Mielke et~al.(2022)Mielke, Szlam, Dinan, and Boureau}]{mielke-etal-2022-reducing}
Sabrina~J. Mielke, Arthur Szlam, Emily Dinan, and Y-Lan Boureau. 2022.
\newblock \href {https://doi.org/10.1162/tacl_a_00494} {Reducing conversational agents{'} overconfidence through linguistic calibration}.
\newblock \emph{Transactions of the Association for Computational Linguistics}, 10:857--872.

\bibitem[{Min et~al.(2023)Min, Krishna, Lyu, Lewis, Yih, Koh, Iyyer, Zettlemoyer, and Hajishirzi}]{min-etal-2023-factscore}
Sewon Min, Kalpesh Krishna, Xinxi Lyu, Mike Lewis, Wen-tau Yih, Pang Koh, Mohit Iyyer, Luke Zettlemoyer, and Hannaneh Hajishirzi. 2023.
\newblock \href {https://doi.org/10.18653/v1/2023.emnlp-main.741} {{FA}ct{S}core: Fine-grained atomic evaluation of factual precision in long form text generation}.
\newblock In \emph{Proceedings of the 2023 Conference on Empirical Methods in Natural Language Processing}, pages 12076--12100, Singapore. Association for Computational Linguistics.

\bibitem[{Murray and Chiang(2018)}]{murray-chiang-2018-correcting}
Kenton Murray and David Chiang. 2018.
\newblock \href {https://doi.org/10.18653/v1/W18-6322} {Correcting length bias in neural machine translation}.
\newblock In \emph{Proceedings of the Third Conference on Machine Translation: Research Papers}, pages 212--223, Brussels, Belgium. Association for Computational Linguistics.

\bibitem[{OpenAI(2022)}]{gpt3.5}
OpenAI. 2022.
\newblock \href {https://openai.com/blog/chatgpt} {Chatgpt blog post}.

\bibitem[{OpenAI(2023)}]{gpt4}
OpenAI. 2023.
\newblock \href {https://arxiv.org/abs/2303.08774} {Gpt-4 technical report}.

\bibitem[{Rivera et~al.(2024)Rivera, Godbout, Rabbany, and Pelrine}]{rivera2024combining}
Mauricio Rivera, Jean-Fran{\c{c}}ois Godbout, Reihaneh Rabbany, and Kellin Pelrine. 2024.
\newblock \href {https://aclanthology.org/2024.uncertainlp-1.12} {Combining confidence elicitation and sample-based methods for uncertainty quantification in misinformation mitigation}.
\newblock In \emph{Proceedings of the 1st Workshop on Uncertainty-Aware NLP (UncertaiNLP 2024)}, pages 114--126, St Julians, Malta. Association for Computational Linguistics.

\bibitem[{Schober et~al.(2018)Schober, Boer, and Schwarte}]{schober2018correlation}
Patrick Schober, Christa Boer, and Lothar~A Schwarte. 2018.
\newblock \href {https://journals.lww.com/anesthesia-analgesia/fulltext/2018/05000/correlation_coefficients__appropriate_use_and.50.aspx} {Correlation coefficients: appropriate use and interpretation}.
\newblock \emph{Anesthesia \& analgesia}, 126(5):1763--1768.

\bibitem[{Stelmakh et~al.(2022)Stelmakh, Luan, Dhingra, and Chang}]{stelmakh-etal-2022-asqa}
Ivan Stelmakh, Yi~Luan, Bhuwan Dhingra, and Ming-Wei Chang. 2022.
\newblock \href {https://doi.org/10.18653/v1/2022.emnlp-main.566} {{ASQA}: Factoid questions meet long-form answers}.
\newblock In \emph{Proceedings of the 2022 Conference on Empirical Methods in Natural Language Processing}, pages 8273--8288, Abu Dhabi, United Arab Emirates. Association for Computational Linguistics.

\bibitem[{Vazhentsev et~al.(2023)Vazhentsev, Tsvigun, Vashurin, Petrakov, Vasilev, Panov, Panchenko, and Shelmanov}]{vazhentsev-etal-2023-efficient}
Artem Vazhentsev, Akim Tsvigun, Roman Vashurin, Sergey Petrakov, Daniil Vasilev, Maxim Panov, Alexander Panchenko, and Artem Shelmanov. 2023.
\newblock \href {https://doi.org/10.18653/v1/2023.findings-acl.93} {Efficient out-of-domain detection for sequence to sequence models}.
\newblock In \emph{Findings of the Association for Computational Linguistics: ACL 2023}, pages 1430--1454, Toronto, Canada. Association for Computational Linguistics.

\bibitem[{Wang et~al.(2023)Wang, Liu, Yue, Tang, Zhang, Jiayang, Yao, Gao, Hu, Qi, Wang, Yang, Wang, Xie, Zhang, and Zhang}]{wang2023survey}
Cunxiang Wang, Xiaoze Liu, Yuanhao Yue, Xiangru Tang, Tianhang Zhang, Cheng Jiayang, Yunzhi Yao, Wenyang Gao, Xuming Hu, Zehan Qi, Yidong Wang, Linyi Yang, Jindong Wang, Xing Xie, Zheng Zhang, and Yue Zhang. 2023.
\newblock \href {http://arxiv.org/abs/2310.07521} {Survey on factuality in large language models: Knowledge, retrieval and domain-specificity}.

\bibitem[{Williams et~al.(2018)Williams, Nangia, and Bowman}]{multinli}
Adina Williams, Nikita Nangia, and Samuel Bowman. 2018.
\newblock \href {https://doi.org/10.18653/v1/N18-1101} {A broad-coverage challenge corpus for sentence understanding through inference}.
\newblock In \emph{Proceedings of the 2018 Conference of the North {A}merican Chapter of the Association for Computational Linguistics: Human Language Technologies, Volume 1 (Long Papers)}, pages 1112--1122, New Orleans, Louisiana. Association for Computational Linguistics.

\bibitem[{Xiao et~al.(2022)Xiao, Liang, Bhatt, Neiswanger, Salakhutdinov, and Morency}]{xiao-etal-2022-uncertainty}
Yuxin Xiao, Paul~Pu Liang, Umang Bhatt, Willie Neiswanger, Ruslan Salakhutdinov, and Louis-Philippe Morency. 2022.
\newblock \href {https://doi.org/10.18653/v1/2022.findings-emnlp.538} {Uncertainty quantification with pre-trained language models: A large-scale empirical analysis}.
\newblock In \emph{Findings of the Association for Computational Linguistics: EMNLP 2022}, pages 7273--7284, Abu Dhabi, United Arab Emirates. Association for Computational Linguistics.

\bibitem[{Xiong et~al.(2023)Xiong, Hu, Lu, Li, Fu, He, and Hooi}]{xiong2023llms}
Miao Xiong, Zhiyuan Hu, Xinyang Lu, Yifei Li, Jie Fu, Junxian He, and Bryan Hooi. 2023.
\newblock \href {http://arxiv.org/abs/2306.13063} {Can llms express their uncertainty? an empirical evaluation of confidence elicitation in llms}.

\bibitem[{Yang et~al.(2023)Yang, Ravikumar, Schmitt-Ulms, Lolla, Demir, Elistratov, Lavaee, Lolla, Ahmadi, Rus, Amini, and Perez}]{yang2023uncertaintyaware}
Qi~Yang, Shreya Ravikumar, Fynn Schmitt-Ulms, Satvik Lolla, Ege Demir, Iaroslav Elistratov, Alex Lavaee, Sadhana Lolla, Elaheh Ahmadi, Daniela Rus, Alexander Amini, and Alejandro Perez. 2023.
\newblock \href {http://arxiv.org/abs/2311.15451} {Uncertainty-aware language modeling for selective question answering}.

\bibitem[{Yao et~al.(2024)Yao, Wu, Guo, Zhou, Gao, Luo, Hou, Fu, and Song}]{yao2024learning}
Yuxuan Yao, Han Wu, Zhijiang Guo, Biyan Zhou, Jiahui Gao, Sichun Luo, Hanxu Hou, Xiaojin Fu, and Linqi Song. 2024.
\newblock \href {https://arxiv.org/abs/2403.19094} {Learning from correctness without prompting makes llm efficient reasoner}.

\bibitem[{Yin et~al.(2023)Yin, Sun, Guo, Wu, Qiu, and Huang}]{yin-etal-2023-large}
Zhangyue Yin, Qiushi Sun, Qipeng Guo, Jiawen Wu, Xipeng Qiu, and Xuanjing Huang. 2023.
\newblock \href {https://doi.org/10.18653/v1/2023.findings-acl.551} {Do large language models know what they don{'}t know?}
\newblock In \emph{Findings of the Association for Computational Linguistics: ACL 2023}, pages 8653--8665, Toronto, Canada. Association for Computational Linguistics.

\bibitem[{Zhang et~al.(2024)Zhang, Guo, and Vlachos}]{zhang2024need}
Caiqi Zhang, Zhijiang Guo, and Andreas Vlachos. 2024.
\newblock \href {http://arxiv.org/abs/2401.15498} {Do we need language-specific fact-checking models? the case of chinese}.

\bibitem[{Zhang et~al.(2020)Zhang, Kishore, Wu, Weinberger, and Artzi}]{bert-score}
Tianyi Zhang, Varsha Kishore, Felix Wu, Kilian~Q. Weinberger, and Yoav Artzi. 2020.
\newblock \href {https://openreview.net/forum?id=SkeHuCVFDr} {Bertscore: Evaluating text generation with {BERT}}.
\newblock In \emph{8th International Conference on Learning Representations, {ICLR} 2020, Addis Ababa, Ethiopia, April 26-30, 2020}. OpenReview.net.

\bibitem[{Zhang et~al.(2023)Zhang, Li, Cui, Cai, Liu, Fu, Huang, Zhao, Zhang, Chen et~al.}]{zhang2023siren}
Yue Zhang, Yafu Li, Leyang Cui, Deng Cai, Lemao Liu, Tingchen Fu, Xinting Huang, Enbo Zhao, Yu~Zhang, Yulong Chen, et~al. 2023.
\newblock \href {https://arxiv.org/abs/2309.01219} {Siren's song in the ai ocean: A survey on hallucination in large language models}.
\newblock \emph{ArXiv preprint}, abs/2309.01219.

\bibitem[{Zhao et~al.(2023)Zhao, Zhou, Li, Tang, Wang, Hou, Min, Zhang, Zhang, Dong, Du, Yang, Chen, Chen, Jiang, Ren, Li, Tang, Liu, Liu, Nie, and Wen}]{LLMSurvey}
Wayne~Xin Zhao, Kun Zhou, Junyi Li, Tianyi Tang, Xiaolei Wang, Yupeng Hou, Yingqian Min, Beichen Zhang, Junjie Zhang, Zican Dong, Yifan Du, Chen Yang, Yushuo Chen, Zhipeng Chen, Jinhao Jiang, Ruiyang Ren, Yifan Li, Xinyu Tang, Zikang Liu, Peiyu Liu, Jian-Yun Nie, and Ji-Rong Wen. 2023.
\newblock \href {https://arxiv.org/abs/2303.18223} {A survey of large language models}.
\newblock \emph{ArXiv preprint}, abs/2303.18223.

\bibitem[{Zheng et~al.(2023)Zheng, Chiang, Sheng, Zhuang, Wu, Zhuang, Lin, Li, Li, Xing, Zhang, Gonzalez, and Stoica}]{zheng2023judging}
Lianmin Zheng, Wei{-}Lin Chiang, Ying Sheng, Siyuan Zhuang, Zhanghao Wu, Yonghao Zhuang, Zi~Lin, Zhuohan Li, Dacheng Li, Eric~P. Xing, Hao Zhang, Joseph~E. Gonzalez, and Ion Stoica. 2023.
\newblock \href {http://papers.nips.cc/paper\_files/paper/2023/hash/91f18a1287b398d378ef22505bf41832-Abstract-Datasets\_and\_Benchmarks.html} {Judging llm-as-a-judge with mt-bench and chatbot arena}.
\newblock In \emph{Advances in Neural Information Processing Systems 36: Annual Conference on Neural Information Processing Systems 2023, NeurIPS 2023, New Orleans, LA, USA, December 10 - 16, 2023}.

\bibitem[{Zhou et~al.(2023)Zhou, Wan, Proleev, Mincu, Chen, Heller, and Roy}]{zhou2023batch}
Han Zhou, Xingchen Wan, Lev Proleev, Diana Mincu, Jilin Chen, Katherine Heller, and Subhrajit Roy. 2023.
\newblock \href {http://arxiv.org/abs/2309.17249} {Batch calibration: Rethinking calibration for in-context learning and prompt engineering}.

\end{thebibliography}

\appendix

\begin{table*}[t!]
\centering
\footnotesize
\begin{tabular}{lcccccc}
\toprule
\textbf{} & GPT-4 & GPT-3.5 & Yi-34B-Chat & Tulu-2-70B & Gemini 1.0 Pro & Vicuna-33B \\
\midrule
\textsc{Luq} & -60.4 & -71.3 & -73.8 & -77.6 & -85.1 & -71.8 \\
\textsc{Luq-Pair} & -61.3 & -72.8 & -76.1 & -80.5 & -86.1 & -72.9 \\
\textsc{Luq-Atomic} & -63.5 & -75.6 & -79.6 & -83.6 & -87.2 & -75.7 \\
\bottomrule
\end{tabular}
\caption{Pearson Correlation Scores between factuality scores and uncertainty scores for \textsc{Luq}'s variations on \textsc{FactScore-bio} dataset.}
\label{tab:luq_variations}
\end{table*}

\section{\textsc{Luq} variations} \label{app:variations}
Table \ref{tab:luq_variations} compares \textsc{Luq}, \textsc{Luq-Pair}, and \textsc{Luq-Atomic}. Our findings indicate that \textsc{Luq-Pair} and \textsc{Luq-Atomic} outperform the original \textsc{Luq} across all models.

The superiority of \textsc{Luq-Pair} stems from its use of shorter premises for NLI (sentence $s'_j$ instead of $r'$), which leads to higher NLI accuracy. However, this improvement comes at the cost of increased computational requirements. For $N$ samples with $M$ sentences each, the original \textsc{Luq} requires $M \times M$ NLI computations, whereas \textsc{Luq-Pair} requires $N \times M^2$ computations.

In \textsc{Luq-Atomic}, we first break down the text into atomic sentences using ChatGPT before proceeding with further steps. The main concern of this variation is about \textbf{evaluation fairness}. Both \textsc{Luq-Atomic} and \textsc{FactScore} use ChatGPT to break long texts into atomic sentences, potentially creating an unfair comparison with other UQ methods that do not involve this step. Further thorough investigation is needed to determine if this approach is universally beneficial, regardless of the atomic fact producer/converter used. This would require a new study and could be a valuable follow-up work. Notably, our original \textsc{Luq} can \textit{still outperform existing baselines without this step}.

Regarding the choice of \textsc{Luq} and its variations, we recommend the following:
\begin{enumerate}
    \item For general purposes and scenarios requiring high efficiency, use \textsc{Luq}.
    \item For cases needing very accurate uncertainty estimation and where time is not a constraint, use \textsc{Luq-Atomic} if the budget allows for API calls. If not, use \textsc{Luq-Pair}.
\end{enumerate}

\section{Dataset Selection} \label{app:dataset_selection}

\subsection{Dataset for Long-form Uncertainty}

As mentioned in Section \ref{sec:dataset}, when selecting the dataset, we considered three main criteria: (1) The dataset should be a long-form QA dataset with relatively lengthy answers. (2) There should be a well-designed and widely-accepted automatic evaluation tool. (3) The questions should be clear, specific, and have definite answers for objective evaluation. 

Evaluating long-form QA is a long-standing challenge, making the last criterion especially important to mitigate factors that could affect evaluation quality. According to \citet{hu2023uncertainty} and \citet{baan2023uncertainty}, uncertainty in NLG systems can be disentangled into three main sources: \textit{input, model, and output}. Due to the intrinsic ambiguity of language and unknown queries, the input itself contains uncertainty \citep{baan2023uncertainty}. To conduct \textbf{a more controlled study}, we focus primarily on \textit{output uncertainty}, assuming all questions are generally answerable and clearly stated. 

Among all the datasets, \textsc{FActScore} advances the field by using LLMs for human-level evaluation, addressing the limitations of traditional metrics like BLEU, ROUGE-L, and BERTScore. Other long-form QA benchmarks fall short in at least one criterion. For example, ELI5 \citep{fan-etal-2019-eli5} questions are very general (e.g., ``How can different animals perceive different colors?”) and can be answered in many ways, making it hard to define objective criteria for a good answer. ASQA \citep{stelmakh-etal-2022-asqa} questions are inherent ambiguous, making it unsuitable for proving the efficiency of an UQ method. 

\subsection{Human Evaluation on \textsc{FActScore}} 

We also engaged human annotators to assess the factuality of the generated passages. Although \citet{min-etal-2023-factscore} conducted comprehensive experiments to demonstrate the effectiveness of the \textsc{FActScore} framework, we perform a sanity check by directly correlating the annotated passage factuality with uncertainty scores. We recruited three students with Master's degrees in Computer Science from our university to conduct the human annotations. We ensured the annotators were not involved in our project and had not discussed it. We used Fleiss' Kappa to measure inter-annotator agreement, achieving a score of 0.793, indicating substantial agreement (close to the ``almost perfect" standard of 0.8-1.0) according to \citet{landis1977measurement}. Annotators are compensated above the local minimum hourly wage standard. The instructions provided to the annotators are listed in Figure \ref{tab:guidelines}.

\begin{table*}[t]
\centering
\footnotesize
\begin{tcolorbox}[title = {Human Annotation Guidelines}]
Your task is to evaluate the veracity of each sentence in the provided passage. It is crucial to carefully assess each statement for accuracy and relevance to the main topic.

\noindent \textbf{Steps to Follow:}
\begin{enumerate}
    \item \textbf{Read the Passage Thoroughly:} Begin by reading the entire passage to grasp the overall context and the main topic.
    \item \textbf{Check Each Sentence:} Examine each sentence individually for accuracy and completeness. Determine if the information is factual and supported by reliable sources, and whether the sentence presents a partial truth or is fully accurate. 
    \item \textbf{Scoring:} Assign each sentence a score based on its accuracy, using a specified range (e.g., 1 to 3). Scores should reflect:
    \begin{itemize}
        \item The sentence is entirely accurate and provides a complete picture. [Highest Score: 3]
        \item The sentence is partially correct but may lack context or omit important details. [Mid-Range Score: 2]
        \item The sentence is largely inaccurate or misleading. [Lowest Score: 1]
    \end{itemize}
    \item \textbf{Relevance:} Flag any sentence that does not contribute to or is off-topic as \textit{Not Relevant}.
\end{enumerate}

\noindent \textbf{Guidelines:}

\begin{itemize}
    \item Use reliable sources (e.g. Wikipedia) to verify factual information, maintaining an impartial stance throughout. 
    \item Keep the passage and your assessments confidential. 
\end{itemize}
\end{tcolorbox}
\caption{Human Annotation Guidelines}
\label{tab:guidelines}
\end{table*}

We randomly selected 50 passages from the responses generated by the Yi-34B-Chat model. We observed a Pearson correlation coefficient of 0.88 between the \textsc{FActScore} factuality score and the human-annotated factuality score. This finding aligns with the results reported by \citet{min-etal-2023-factscore}, demonstrating that \textsc{FActScore} is a reliable tool in our experiments. Table \ref{tab:human_annotation} compares the results of different UQ methods with those obtained using \textsc{FActScore} and human annotation. 

\section{\textsc{FActScore-Dis}} \label{app:factscore-dis}

\begin{table}[]
\footnotesize
\centering
\renewcommand\arraystretch{1.5}
\tabcolsep=1.5mm
\begin{tabular}{l *{4}{r}}
\toprule

\multirow{2}{*}{Methods} & \multicolumn{2}{c}{FactScore} & \multicolumn{2}{c}{Human} \\

 & \multicolumn{1}{c}{PCC} & \multicolumn{1}{c}{SCC} & \multicolumn{1}{c}{PCC} & \multicolumn{1}{c}{SCC} \\

\midrule
LexSimilarity & -67.3 & -66.4 & -65.6 & -64.0  \\
Eccentricity & -26.3 & -25.5 & -22.6 & -25.1  \\
NumSemSets & -26.4 & -26.9 & -24.3 & -23.5  \\
EigValLaplacian & -45.8 & -43.9 & -43.4 & -42.7 \\
DegMat & -38.9 & -39.7 & -36.8 & -31.6 \\
SelfCheckNLI & -68.5 & -67.3 & -66.1 & -69.2 \\
\textsc{Luq} & \textbf{-72.7} & \textbf{-71.4} & \textbf{-69.0} & \textbf{-68.3} \\

\bottomrule
\end{tabular}
\caption{Pearson and Spearman correlation coefficients (expressed as percentages) between different factuality scores and various UQ methods on the \textbf{FactScore-Bio} dataset using Yi-34B-Chat.}
\label{tab:human_annotation}
\vspace{-3mm}
\end{table}

To demonstrate the generalization of our proposed \textsc{Luq} model across various domains, we create a new dataset adopting the methodology used to construct the original \textsc{FActScore} dataset for the disease entities. To differentiate, we refer the original dataset as \textsc{FActScore-Bio} and the new dataset as \textsc{FActScore-Dis}. The detailed information of \textsc{FActScore-Dis} dataset is as follows: 

\paragraph{Data Collection} Following \textsc{FActScore}-Bio, we use Wikipedia as our main knowledge source. We first select all the diseases names using the following SPARQL codes calling the wiki API. We then removed those diseases with empty Wikipedia pages.  

Following \textsc{FActScore-Bio}, we utilized Wikipedia as our primary knowledge source. Initially, we extracted all disease names using the following SPARQL queries to call the Wikidata API. Subsequently, we removed those diseases with empty Wikipedia pages. 

\definecolor{codegreen}{rgb}{0,0.6,0}
\definecolor{codegray}{rgb}{0.5,0.5,0.5}
\definecolor{codepurple}{rgb}{0.58,0,0.82}
\definecolor{backcolour}{rgb}{0.95,0.95,0.92}

\lstdefinestyle{mystyle}{
    backgroundcolor=\color{backcolour},   
    commentstyle=\color{codegreen},
    keywordstyle=\color{magenta},
    numberstyle=\tiny\color{codegray},
    stringstyle=\color{codepurple},
    basicstyle=\ttfamily\footnotesize,
    breakatwhitespace=false,         
    breaklines=true,                 
    captionpos=b,                    
    keepspaces=true,                                
    showspaces=false,                
    showstringspaces=false,
    showtabs=false,                  
    tabsize=2
}

\lstset{style=mystyle}

\begin{lstlisting}[language=SQL]
SELECT ?item ?itemLabel WHERE {
  ?item wdt:P31 wd:Q112193867. # is an instance of class of diseases
  SERVICE wikibase:label { bd:serviceParam wikibase:language "[AUTO_LANGUAGE],en". }
}
\end{lstlisting}

\paragraph{Frequency} 

For each entity retrieved, we adhere to the methodology described by \citet{min-etal-2023-factscore} to assign a frequency label ranging from ``Very Rare" to ``Very Frequent" based on an entity's pageviews. It's crucial to acknowledge that in the context of diseases, the number of diagnosed cases is commonly used as a metric. However, we opted not to use this metric because our goal is to simulate the distribution of these diseases within the training corpus of LLMs. Relying solely on diagnosed case numbers may underrepresent the prominence of a disease within the corpus. Diseases like Amyotrophic Lateral Sclerosis (ALS), despite their low incidence rate in the population, attract significant global interest and impact. As a result, LLMs may demonstrate extensive knowledge about such diseases, reflecting their visibility in the data on which they are trained, rather than their actual morbidity rates.

After determining the frequencies, we sampled 36 disease entities for each category, amassing a total of 180 data points. Subsequently, we conducted a human evaluation to validate the selected data points, replacing any that were deemed unsuitable with diseases that were more clearly defined and well-documented. Several examples from the dataset are showcased in Table \ref{tab:FrequencyOfDiseases}.

\begin{table}[ht!]
\centering
\footnotesize
\begin{tabular}{lll}
\toprule
\textbf{Frequency} & \textbf{Wikidata ID} & \textbf{Disease Name} \\
\midrule
\multicolumn{3}{l}{\textbf{Very Freq}} \\
 & Q8071861 & Zika fever \\
 & Q12199 & HIV/AIDS \\
 & Q12152 & myocardial infarction \\
 & Q12206 & diabetes \\
 & Q12204 & tuberculosis \\
\addlinespace
\multicolumn{3}{l}{\textbf{Freq}} \\
 & Q154874 & yellow fever \\
 & Q188638 & mood disorder \\
 & Q159701 & glaucoma \\
 & Q1138580 & Ewing sarcoma \\
 & Q209369 & Hodgkin lymphoma \\
\addlinespace
\multicolumn{3}{l}{\textbf{Medium}} \\
 & Q5134736 & cloacal exstrophy \\
 & Q247978 & anisometropia \\
 & Q2373361 & tree nut allergy \\
 & Q778731 & pyuria \\
 & Q7900433 & urethral syndrome \\
\addlinespace
\multicolumn{3}{l}{\textbf{Rare}} \\
 & Q220322 & agnosia \\
 & Q2735907 & cutis laxa \\
 & Q500695 & retinoblastoma \\
 & Q627625 & histoplasmosis \\
 & Q1347729 & Epstein syndrome \\
\addlinespace
\multicolumn{3}{l}{\textbf{Very Rare}} \\
 & Q21505502 & spina bifida \\
 & Q1862031 & pinguecula \\
 & Q1361850 & patulous eustachian tube \\
 & Q4667534 & leiomyoma \\
 & Q595010 & hypertrichosis \\
\bottomrule
\end{tabular}
\caption{Frequency Categories of Diseases}
\label{tab:FrequencyOfDiseases}
\end{table}







\section{Experiment Setup} \label{app:setup}

For GPT-4 and GPT-3.5, we use the OpenAI API, with specific version \texttt{gpt-4-turbo-0125-preview} and \texttt{gpt-3.5-turbo-0613}. For Gemini 1.0 Pro, we call the API for developers. For Yi-34B-Chat, Tulu-2-70B (\texttt{tulu-2-dpo-70b}), and Vicuna-33B (\texttt{vicuna-33b-v1.3}), we use them off-the-shelf and only for inference. The temperature is set to 0.7. We run our uncertainty measurement experiments on A100-SXM-80GB GPUs. For our experiments, we use the following prompt: 

\begin{mdframed}[backgroundcolor=gray!20, linecolor=black, linewidth=1pt, roundcorner=5pt]
\textit{Tell me a short bio of the person <entity>. Begin with their birth, significant life events, achievements, and contributions. Include their education, career milestones, any notable awards or recognitions received, and their impact on their field or society.  Ensure the biography is concise, factual, and engaging, covering key aspects of their life and work.}

\end{mdframed}

From the esitimation of \citet{min-etal-2023-factscore}, running \textsc{FActScore} costs about \$1 of the API cost per 100 sentences. For instance, for 100 generations, each with 5 sentences on average, it costs \$5 in total. 

\section{Baselines} \label{app:baselines}

We mainly use the library LM-Polygraph \citep{fadeeva-etal-2023-lm} for the UQ methods. Here we provide a brief introduction for each method: 

\noindent
\textbf{LexicalSimilarity} \citep{fomicheva-etal-2020-unsupervised}: it computes the similarity between two phrases using metrics like ROUGE scores and BLEU. For our experiment, we utilize BERTScore \citep{bert-score} to enhance performance, computing the average similarity score with other answers.

\noindent
\textbf{NumSemSets} \citep{lin2023generating}: it clusters semantically equivalent answers into the same sets. Initially, the number of semantic sets equals the total number of generated answers. Then it sequentially examines responses, making pairwise comparisons between them, and combines different answers. One of the limitation of this method is that the uncertainty score $U_{N u m S e m S e t s}$ can only take integer values. EigValLaplacian is therefore designed to overcome this problem. 

\noindent
\textbf{EigValLaplacian} \citep{lin2023generating}: For a similarity matrix $S$, it calculates the
Normalized Graph Laplacian of $S$ using $L=I-D^{-\frac{1}{2}} S D^{-\frac{1}{2}}$, where $D$ is a diagonal matrix and $D_{i i}=\sum_{j=1}^m S_{i j}$ ($m$ is the number of responses). Consequently, the uncertainty score is defined as $U_{E i g V}=\sum_{k=1}^m \max \left(0,1-\lambda_k\right)$. This value is a continuous analogue of $U_{N u m S e m S e t s}$. In extreme case if adjacency matrix $S$ is binary these two measures will coincide. 

\noindent
\textbf{DegMat} \citep{lin2023generating}: it is based on the idea that the total uncertainty of the answers might be measured as a corrected trace of the diagonal matrix $D$. This is because elements on the diagonal of matrix D are sums of similarities between the given answer and other answers. We thus define uncertainty estimate $
U_{\text {Deg }}(x)=\operatorname{trace}(m-D) / m^2 \quad 
$.

\noindent
\textbf{Eccentricity} \citep{lin2023generating}: A drawback of previously considered methods is the limited knowledge of the actual embedding space for the different answers since we only have measures of their similarities. The graph Laplacian, however, can provide us with coordinates for the responses. Denote $\mathbf{u}_1, \ldots, \mathbf{u}_k \in \mathbb{R}^m$ as the eigenvectors of $L$ that correspond to $k$ smallest eigenvalues. We can efficiently construct an informative embedding $\mathbf{v}_j=\left[\mathbf{u}_{1, j}, \ldots, \mathbf{u}_{k, j}\right]$ for an answer $\mathbf{y}_j$. Then it uses the average distance from center as the uncertainty measure, defined as : $U_{Ecc}=$ $\left\|\left[\tilde{\mathbf{v}}_1^T, \ldots, \tilde{\mathbf{v}}_m^T\right]\right\|_2$, where $\tilde{\mathbf{v}}_j=\mathbf{v}_j-\frac{1}{m} \sum_{\ell=1}^m \mathbf{v}_{\ell}$.

\paragraph{SelfCheckNLI} \cite{manakul-etal-2023-selfcheckgpt}: As defined in Section \ref{sec:background}, SelfCheckNLI primarily functions as a confidence measurement tool, calculating the similarity exclusively between the primary response $r_a$ and the other generated samples. Distinctively, it evaluates $\mathcal{P}(\text{contradict} \mid s, r')$ and focuses solely on $C(x, r_a)$.

\section{Response Statistics} 

\paragraph{Average Response Length.} In our study, we define ``long text” as the model’s output consisting of at least 100 words. For our experiments, the texts are even longer, averaging more than 200 words. The average word count is calculated and shown in the following table. In contrast, commonly used datasets for existing UQ methods have much shorter texts, with an average of 1.95 words for Trivia QA dataset and 3.37 words for Natural Questions dataset. 

\begin{table}[h]
\footnotesize
\begin{tabular}{lc}
\toprule
\textbf{Model}     & \textbf{Avg. Word Count per Response} \\
\midrule
GPT-4-0125         & 264                                      \\
GPT-3.5-turbo-0613 & 239                                      \\
Gemini Pro 1.0     & 223                                      \\
Yi-34B-Chat        & 307                                      \\
Tulu-2-70B         & 267                                      \\
Vicuna-33B         & 206                                      \\
\bottomrule
\end{tabular}
\caption{The average response length for each LLM.}
\end{table}

\begin{figure}[ht!]
    \centering
    \includegraphics[width=0.9\columnwidth]{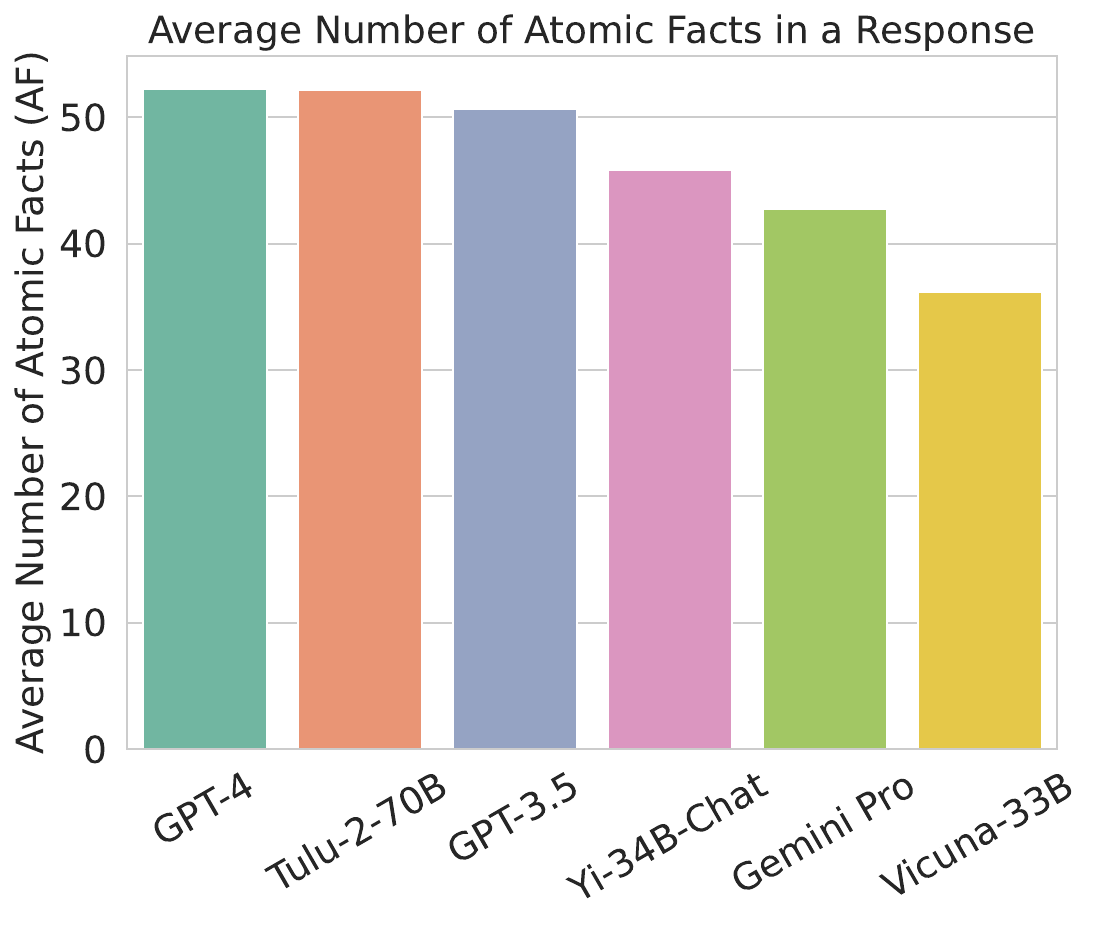}
    \caption{Average number of atomic Facts (AF) in a response for each model.}
    \label{fig:number_of_facts}
\end{figure}

\paragraph{Number of Facts in a Response}
Figure \ref{fig:number_of_facts} shows the average atomic facts provided by various AI models for the \textsc{FActScore} dataset. GPT-4 has the highest average number of atomic facts at 52.24, indicating it provides the most detailed factual responses. Tulu-2-70B follows with an average of 52.17, nearly matching GPT-4 in factual details. GPT-3.5 has an AF of 50.67, showing it also delivers a high level of factual details in its responses. Yi-34B-Chat and Gemini 1.0 Pro have comparatively lower averages, at 45.80 and 42.72 respectively. Vicuna-33B has the lowest AF at 36.20, indicating it offers the least amount of factual information in its responses. Generally, these models provide similar number of atomic facts in their responses.

\section{\textsc{Luq-Ensemble}} \label{app:ensemble}

In this section, we discuss more about the novelty, motivation, and effectiveness of our \textsc{Luq-Ensemble} method. 

\rparagraph{The Motivation of \textsc{Luq-Ensemble}} After getting multiple answers from different LLMs, the challenge is now to \textit{choose which one as the final output}. Traditional aggregation methods like majority vote and weighted vote are ineffective for long text generation because finding the majority answer is difficult when all the responses are somewhat different, and methods like boosting or bagging require additional training. Our uncertainty measure thus serves as an effective indicator for model ensembling. 

\rparagraph{The Effectiveness of \textsc{Luq-Ensemble}} Meanwhile, it is not guaranteed that ensembling will lead to better performance. It is true that the uncertainty scores will by definition decrease (as we select the model with the least uncertainty as the final response), but the factuality score may not. The effectiveness of \textsc{Luq-Ensemble} largely relies on the reliability of the UQ method.

Table \ref{tab:different_ensembles} compares the effectiveness of using \textsc{Luq} as the ensemble indicator with other methods. The results show that ensembling \textit{does not inherently improve performance}. With a poor UQ method (such as Ecc), the ensemble factuality score can be lower than that of its highest component (47.2 vs 43.3). In contrast, using \textsc{Luq} as the indicator for ensembling yields the best overall performance.

\begin{table}[h]
\centering
\begin{tabular}{lcc}
\toprule
\textbf{} & \textbf{PFS} & \textbf{PUS} \\
\hline
\rowcolor{gray!30}
\multicolumn{3}{l}{\textbf{Individual Results}} \\
Tulu-2-70B & 47.2 & 55.8  \\
Gemini 1.0 Pro & 42.7 & 62.2 \\
Vicuna-33B & 42.5 & 58.1 \\

\rowcolor{gray!30}
\multicolumn{3}{l}{\textbf{Different Ensemble Methods}} \\
\textsc{Ecc-Ensemble} & 43.4 & \textbf{35.5} \\
\textsc{LexSim-Ensemble} & 47.6 & 39.8 \\
\textsc{SelfCheck-Ensemble} & 49.3 & 46.7 \\
\textsc{Luq-Ensemble} & \textbf{52.8} & 45.8 \\
\bottomrule
\end{tabular}
\caption{Penalised factuality score (PFS) and Penalised uncertainty score (PUS) for individual models and ensembles with different UQ methods.}
\label{tab:different_ensembles}
\end{table}

\section{Selective QA Strategy} \label{app:selective_qa}

When implementing a selective answering strategy in practical applications, it is essential for practitioners to tailor the uncertainty thresholds to the specific models and tasks at hand. In our experiment, as shown in Table \ref{tab:factscores}, we find that GPT-4 tends to refuse to answer around 15\% of the questions. To simulate a GPT-4-like answering strategy, for each model, we set different thresholds to ensure they refuse to answer between 0 and 15\% of the questions. Our experiments indicate that different LLMs may have varying average absolute uncertainty values, making a universal uncertainty threshold unsuitable for all models. Additionally, the inherent nature of the tasks may influence practitioners' decisions to make the model more conservative or more willing to attempt answering users' questions. 

We advise practitioners to implement selective QA strategies using the following practical steps:
\begin{itemize}
    \item Collect a representative set of questions/queries that closely mimic real-world usage scenarios.
    \item Obtain responses from the LLMs for these questions and apply UQ methods (e.g., \textsc{Luq}) to get uncertainty scores.
    \item Establish thresholds tailored to the specific task and the practitioners' goals, selecting either cautious (lower threshold) or lenient (higher threshold) settings as needed.
    \item Develop clear strategies for handling high uncertainty, such as refusing to answer, requesting clarification, or using alternative approaches.
\end{itemize}

\section{Ablation Study} \label{sec:ablation}

\begin{figure}[t!]
    \centering
    \begin{subfigure}[b]{0.4\textwidth}
        \includegraphics[width=\columnwidth]{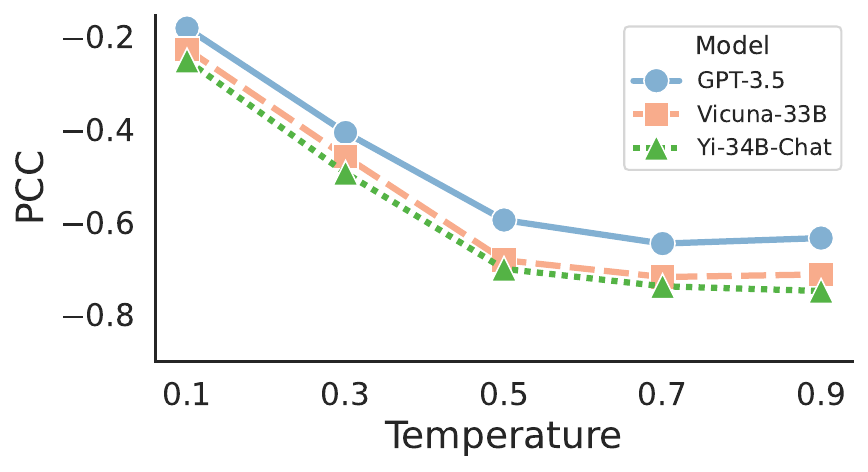}
        \label{fig:temperature}
    \end{subfigure}
    \hfill 
    \begin{subfigure}[b]{0.4\textwidth}
        \includegraphics[width=\columnwidth]{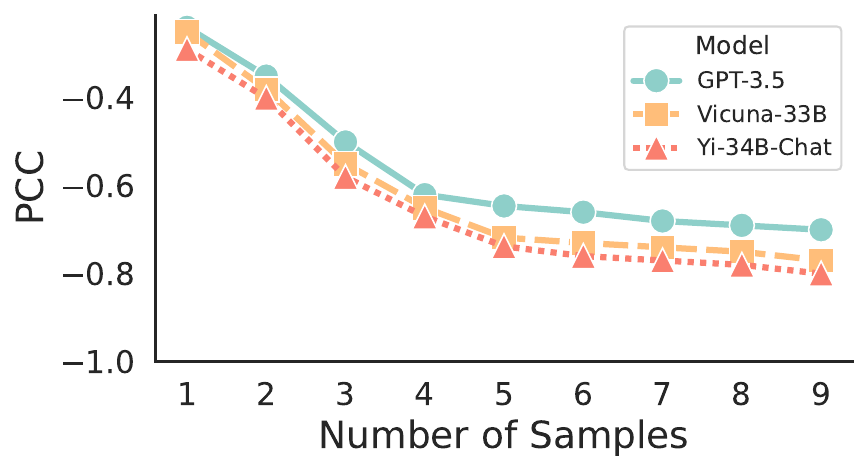}
        \label{fig:number_samples}
        \vspace{-5mm}
    \end{subfigure}
    \caption{The effect of different temperatures (upper) and the number of samples (lower) on the PCC with \textsc{Luq}.}
    \label{fig:combined}
    \vspace{-3mm}
\end{figure}

\paragraph{Temperature} 
As the diversity of content generated by LLMs may be influenced by the temperature setting, we adjust the temperature to test the robustness of our methods. Due to limitations in computational resources and API budget constraints, we selected GPT-3.5, Yi-34B-Chat, and Vicuna-33B for our experiments (refer to Figure \ref{fig:combined}). Our findings indicate that a lower temperature leads to a weaker correlation score, likely because the generated responses are more uniform, providing limited information for the self-consistency test. As the temperature increases, we observe a strengthening in correlation. However, beyond a certain point, further increases in temperature lead to diminishing improvements and can even result in a weaker correlation. We hypothesize that excessively diverse responses may complicate the NLI process, as a greater number of sentences fail to be supported by other samples. 

\paragraph{Number of Samples}
Previous research on short answer generation \citep{kuhn2022semantic, lin2023generating} has demonstrated that an increase in the number of samples correlates with enhanced performance. We investigate whether it also applies to long-text generation and find that with more samples, \textsc{Luq} shows better performance and PCC scores, which corroborates with previous observations in short-text generation, as depicted in Figure\ref{fig:combined}. Providing a greater number of samples enables the NLI process to predict sentence factuality with higher accuracy. However, a notable drawback of increasing the sample size is the associated rise in computational costs. 

\newpage
\onecolumn
\section{Case Study} \label{app:case_study}

In this section, we present two case studies illustrating the performance of \textsc{Luq}. For simplicity, we only show three samples for each question using Yi-34B-Chat. \textit{In both cases, the baseline UQ methods indicate low uncertainty.} However, the first case is with higher factuality, while the second case includes many non-factual information. Unlike baseline UQ methods that can only calculate similarity at the passage level, our method captures more nuanced differences.

\begin{figure*}[h]
    \centering
    \includegraphics[width=.99\textwidth]{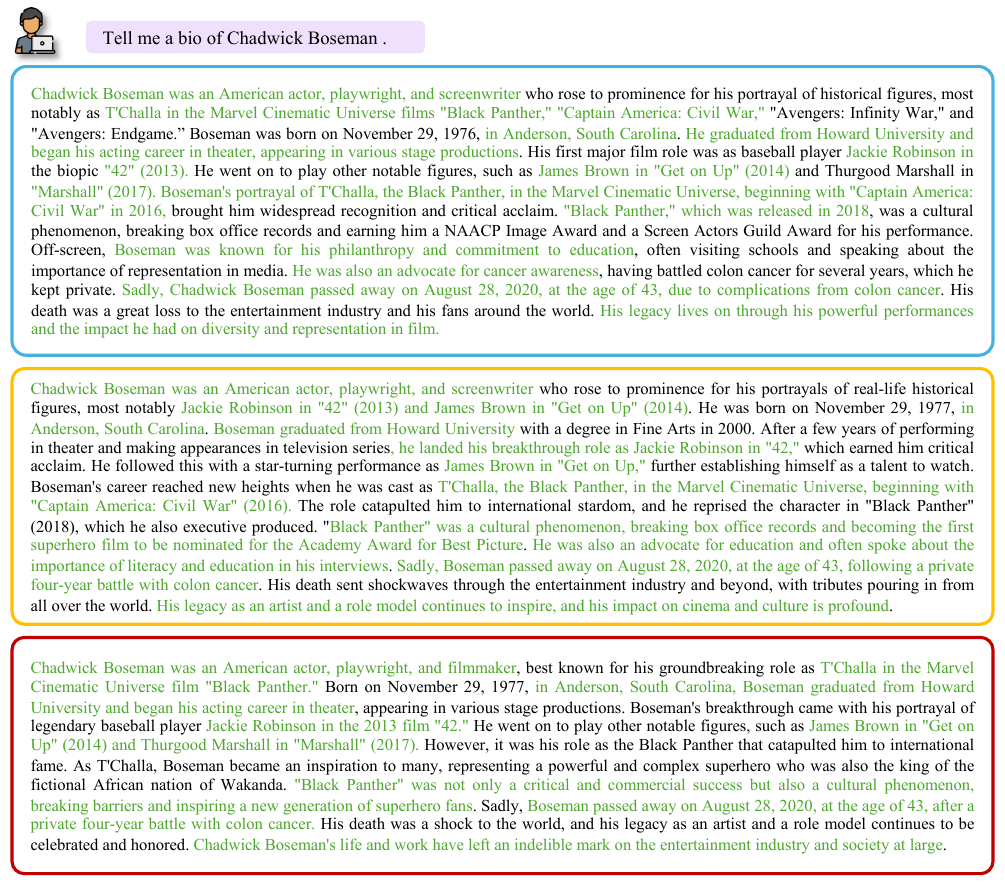}
    \caption{Example with \textbf{low uncertainty} by \textsc{Luq}. Sentences that can be cross-validated by other samples are shown in green.}
    \label{fig:consistent_example}
\end{figure*}

In the first case study, \textsc{Luq} demonstrates lower uncertainty. As shown in Figure \ref{fig:consistent_example}, we use green to represent sentences that can be cross-validated by other samples. It is evident that most sentences are consistent across different samples, thus leading to lower uncertainty.

\begin{figure*}[h]
    \centering
    \includegraphics[width=.99\textwidth]{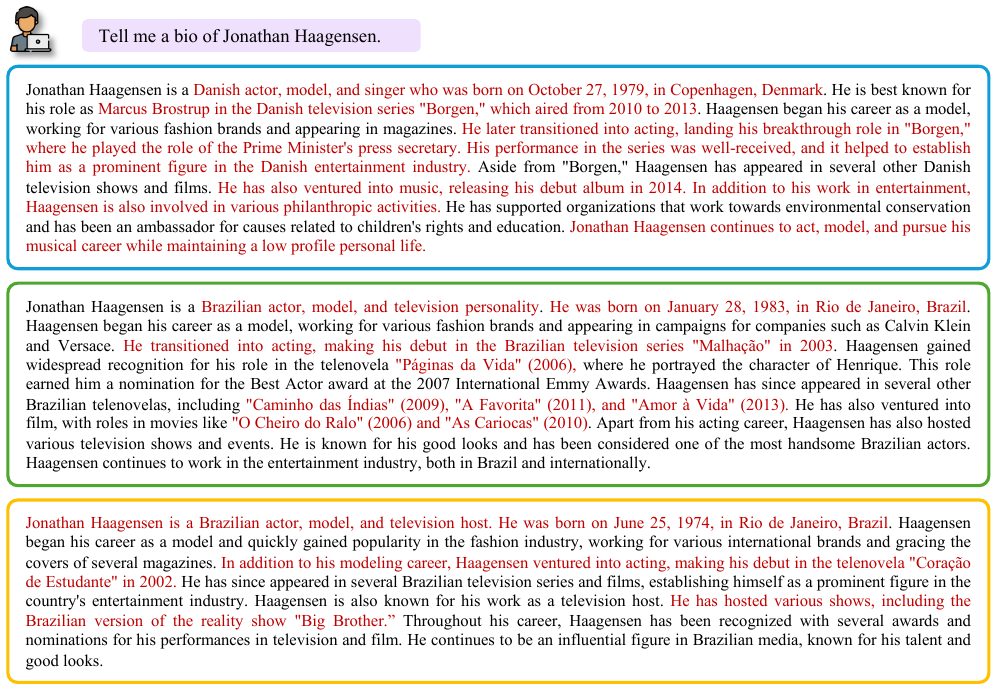}
    \caption{Example with \textbf{high uncertainty} by \textsc{Luq}. Contradictory information is highlighted in red.}
    \label{fig:non_consistent_example}
\end{figure*}

In the second case study, we showcase an example with higher \textsc{Luq} uncertainty. In Figure \ref{fig:non_consistent_example}, red is used to indicate contradictory information. It is apparent that the samples are not consistent with each other. The samples even state different nationalities of Jonathan Haagensen.

\end{document}